\documentclass[sn-apa]{sn-jnl}

\usepackage{graphicx}%
\usepackage{multirow}%
\usepackage{amsmath,amssymb,amsfonts}%
\usepackage{amsthm}%
\usepackage{mathrsfs}%
\usepackage[title]{appendix}%
\usepackage{xcolor}%
\usepackage{textcomp}%
\usepackage{manyfoot}%
\usepackage{booktabs}%
\usepackage{algorithm}%
\usepackage{algorithmicx}%
\usepackage{algpseudocode}%
\usepackage{listings}%
\usepackage{appendix}
\usepackage[T1]{fontenc}

\theoremstyle{thmstyleone}%
\theoremstyle{thmstyletwo}%

\theoremstyle{thmstylethree}%

\newif\ifarxiv
\arxivtrue

\begin{document}

\title{Expandable Decision‑Making States for Multi‑Agent Deep Reinforcement Learning in Soccer Tactical Analysis}

\ifarxiv
\author[1]{\fnm{Kenjiro} \sur{Ide}}\email{ide.kenjiro@g.sp.m.is.nagoya-u.ac.jp}

\author[2]{\fnm{Taiga} \sur{Someya}}\email{tsomeya@g.ecc.u-tokyo.ac.jp}

\author[3]{\fnm{Kohei} \sur{Kawaguchi}}\email{k.kawaguchi@ust.hk}

\author*[1]{\fnm{Keisuke} \sur{Fujii}}\email{fujii@i.nagoya-u.ac.jp}

\affil[1]{\orgdiv{Graduate School of Informatics}, \orgname{Nagoya University}, \orgaddress{\city{Nagoya}, \country{Japan}}}

\affil[2]{\orgdiv{Graduate School of Arts and Sciences}, \orgname{The University of Tokyo}, \orgaddress{\city{Tokyo}, \country{Japan}}}

\affil[3]{\orgdiv{Graduate School of Economics}, \orgname{Hong Kong University of Science and Technology}, \orgaddress{\city{Hong Kong}, \country{China}}}

\else
Anonymous
\fi 


\abstract{
Invasion team sports such as soccer produce a high-dimensional, strongly coupled state space as many players continuously interact on a shared field, challenging quantitative tactical analysis. Traditional rule-based analyses are intuitive, while modern predictive machine learning models often perform pattern-matching without explicit agent representations.
The problem we address is how to build player-level agent models from data, whose learned values and policies are both tactically interpretable and robust across heterogeneous data sources.
Here, we propose Expandable Decision-Making States (EDMS), a semantically enriched state representation that augments raw positions and velocities with relational variables (e.g., scoring of space, pass, and score), combined with an action-masking scheme that gives on-ball and off-ball agents distinct decision sets.
Compared to prior work, EDMS maps learned value functions and action policies to human-interpretable tactical concepts (e.g., marking pressure, passing lanes, ball accessibility) instead of raw coordinate features, and aligns agent choices with the rules of play.
In the experiments, EDMS with action masking consistently reduced both action-prediction loss and temporal-difference (TD) error compared to the baseline.
Qualitative case studies and Q-value visualizations further indicate that EDMS highlights high-risk, high-reward tactical patterns (e.g., fast counterattacks and defensive breakthroughs).
We also integrated our approach into an open-source library and demonstrated compatibility with multiple commercial and open datasets, enabling cross-provider evaluation and reproducible experiments.

}





\maketitle

\section{Introduction}
\label{sec:introduction}
In invasion team sports such as soccer, 22 players interact continuously on a two-dimensional field, which generates a state space that is both high-dimensional and strongly coupled. Early analytical systems coped with this complexity by invoking handcrafted geometric or kinematic rules to estimate ball access or territorial dominance (e.g., \cite{taki1996development,Link2016}). Although these rule-based models remain intuitive for coaches, they cover only narrowly defined situations, such as analyzing shots, defensive pressure, or crosses, and do not scale such that they evaluate all actions of all players. 
Recently, a large body of work in machine learning has evaluated player movements by predicting discrete events (e.g., passes, shots in \cite{Decroos19,simpson2022seq2event}) or by forecasting future trajectories \citep{le2017coordinated,teranishi2022evaluation,fujii2024decentralized}. While powerful for prediction and useful for many practical analytics tasks, these methods mainly perform pattern-matching from inputs to outcomes rather than explicitly modeling each player as an autonomous decision-making agent with internal goals and policies.
More recent studies have adopted deep reinforcement learning (DRL) to infer value functions directly from event and tracking data (e.g., \cite{rahimian2022beyond,nakahara2023action}).
DRL methods can directly learn temporally-extended value functions from raw event and tracking streams, capturing long-horizon strategic effects of actions and enabling counterfactual, policy-oriented evaluation that goes beyond pointwise prediction. 

Existing player-level DRL models by \cite{nakahara2023action} and its open-source reimplementation in OpenSTARLab \citep{yeung2025openstarlab}, share key limitations that constrain tactical insight because both simply encode match context solely with the Cartesian coordinates and velocities of all players and the ball, which omits relational cues such as marking pressure or passing lanes. They also apply a single, identical discrete action set to every agent, so off-ball players are permitted to select ball-centric options like pass or shot, despite never possessing the ball. These two mismatches may degrade the interpretability of the estimated action values, increase action loss and temporal-difference error in RL (i.e., decrease reward prediction).
Moreover, their evaluation is confined to single-domain case studies, and there is a lack of publicly released implementations and cross-domain benchmarking, i.e., shared datasets and reproducible pipelines across leagues, seasons, or providers.

In this paper, we propose Expandable Decision-Making States (EDMS). This semantically rich yet computationally efficient representation augments raw positions and velocities with relational descriptors such as a space score to quantify the space occupied by players, the time to reach a player to measure defensive pressure, and the shot score to evaluate scoring opportunities. In addition, to enhance both action prediction and value stability, we introduce an action-masking scheme \citep{silver2017mastering,zahavy2018learn} that separates the action sets of on-ball and off-ball agents, so only the ball carrier may choose pass, dribble, or shot, while teammates decide among support moves. We integrate the entire pipeline into the library called OpenSTARLab RLearn \citep{yeung2025openstarlab} and demonstrate its applicability to other public datasets.

Our main contributions are as follows. (1) We propose a player-level DRL framework that integrates the EDMS, enabling learned value functions and action policies to be mapped to human-interpretable tactical concepts (e.g., marking pressure, passing lanes, ball accessibility).
(2) We introduce an action-masking methodology that assigns distinct decision spaces to on-ball and off-ball agents, which reduces both action-prediction loss and temporal-difference (TD) error.
(3) We conduct a comprehensive evaluation using multiple data providers: proprietary J.League data supplied by DataStadium\footnote{\url{https://datastadium.co.jp/en}}, commercial provider datasets used for LaLiga by StatsBomb\footnote{\url{https://www.hudl.com/en_gb/products/statsbomb}} and SkillCorner\footnote{\url{https://skillcorner.com}}, and openly released benchmarks including PFF FC’s 2022 World Cup tracking/event data\footnote{\url{https://www.blog.fc.pff.com/blog/pff-fc-release-2022-world-cup-data}} and the SoccerNet Game State Reconstruction (GSR) resources \citep{somers2024soccernet}\footnote{\url{https://github.com/SoccerNet/sn-gamestate}}. We also release our implementation as open-source so that reproductions and cross-provider benchmarking are directly possible.
The remainder of the paper is organized as follows: Section 2 surveys related work, Section 3 details the proposed model, Section 4 reports experimental results, Section 5 discusses multi-dataset integration, and Section 6 concludes this study.

\section{Related work}
\label{sec:related_work}
This section introduces existing research on multi-agent RL studies, soccer play evaluation studies, and sports RL studies relevant to team sports.

\subsection{Multi-agent RL studies}
Multi-Agent Reinforcement Learning (MARL) is a framework for modeling the interactions of multiple agents. MARL research can be broadly divided into two approaches: the forward approach, which seeks optimal behavior through simulations, and the backward approach, which analyzes the intentions behind actions from real-world data. In the forward approach, a foundational method is Independent Q-Learning (IQL) \citep{nijssen2006iql}, where each agent learns independently. To improve coordination and handle the complexities of multi-agent settings, the current mainstream forward approach is ``Centralized Training with Decentralized Execution'' (CTDE). In CTDE, information from all agents is used during training to ensure stable learning, while during execution, each agent makes decisions based solely on its own observations. Within CTDE, value decomposition methods are effective for cooperative tasks. These methods represent the team's overall value function as a combination of each agent's value function, with VDN \citep{sunehag2017value}, QMix \citep{rashid2020monotonic}, and QTRAN \citep{son2019qtran} being notable examples. On the other hand, Actor-Critic methods are powerful for tasks with a mix of cooperation and competition, with MADDPG \citep{lowe2017multi} as a representative model. It achieves stable learning by using a critic that incorporates information about the action of other agents during training.

The backward approach, in contrast, aims to estimate a player's behavioral model from real-world data. This category includes Offline Reinforcement Learning, which learning policies from a fixed dataset without further interaction, and Inverse Reinforcement Learning \citep{ng2000algorithms}, which infers the reward function from expert demonstrations. Imitation Learning \citep{le2017coordinated}, which directly learns a policy from expert data, is also a key method in this approach. The specific applications of these methods to sports analytics will be discussed in detail in Section \ref{sec:sports_RL_studies}.

In this study, to model the behavior of actual players from data, we adopt an independent Offline RL method from the perspective of the backward approach.

\subsection{Soccer play evaluation studies}
Research on play evaluation in soccer analytics aims to quantify the impact of individual plays on match outcomes, and the approaches are diverse. Early studies focused on shots, leading to the widespread adoption of the Expected Goals (xG) model, which calculates the probability of a goal from a given shot's position and context. However, since xG cannot evaluate non-shot plays, a more comprehensive evaluation framework has been sought.

One such approach is the evaluation of all on-ball actions. VAEP (Valuing actions by Estimating Probabilities) \citep{decroos2019actions} proposed a framework that calculates the value of every action, including passes and dribbles, based on how they change the subsequent probabilities of scoring and conceding goals. A similar approach has been applied to pass evaluation, with studies conducted to objectively measure the risk (probability of interception) and reward (probability of advancing the attack) of a pass \citep{power2017not}. These methods have made it possible to evaluate all plays on a unified scale by breaking down the sparse reward of a goal into a denser contribution to scoring probability.

The advent of tracking data (full positional information of players and the ball) has enabled the evaluation of off-ball movements and spatial value. The Expected Possession Value (EPV) \citep{fernandez2019decomposing} model uses deep learning to estimate the value of possession at any given moment on the pitch (the probability of it leading to a future goal), achieving situational assessment that includes off-ball movements. Additionally, Expected Threat (xT) \citep{singh2018introducing} introduced the concept of defining the threat level of each area on the pitch and evaluating the act of moving the ball into more advantageous areas.

Furthermore, research is advancing to quantify not only individual plays but also team-wide tactical movements. In addition to classic studies that use Voronoi diagrams formed by player positions to assess a team's spatial control \citep{taki1996development, fujimura2005geometric}, other research has dynamically modeled control by considering player velocity and acceleration \citep{brefeld2019probabilistic}, or measured how effectively space is created for teammates through player movement \citep{fernandez2018wide}. Defense-specific research has also evolved, with the proposal of GVDEP \citep{umemoto2022location}, which evaluates how much a team's defensive formation reduces the probability of conceding a goal, and a framework that assesses a team's defensive strength based on probabilistic models of ball recovery and being attacked \citep{toda2022evaluation}.

Recent advances in deep learning have enabled an approach that treats play evaluation as a prediction task. Transformer-based models can predict with high accuracy what type of play will occur next \citep{simpson2022seq2event}, as well as when and where \citep{yeung2023transformer}. In addition to event prediction, trajectory prediction \citep{le2017coordinated,fujii2024decentralized} and its application to play evaluations \citep{teranishi2022evaluation,fujii2024estimating} have been examined. Moreover, tactical support systems like TacticAI \citep{wang2024tacticai} have also been developed, where an AI learns tactical patterns and recommends the most effective plays in specific situations, such as corner kicks. 
While these predictive models are powerful for post-hoc evaluation and forecasting future developments, their analysis is primarily based on pattern matching from input to outcome. In contrast, by explicitly modeling each player as an autonomous decision-making agent with internal goals and policies, our research goes beyond merely reproducing observed patterns. This approach makes it possible to capture the tactical intent behind actions and to evaluate the quality of play from multiple perspectives.

\subsection{Sports RL studies}
\label{sec:sports_RL_studies}
RL has been applied to the analysis of not only soccer but also various other team and competitive sports.
In the forward approach, research in soccer was significantly accelerated by the advent of an open-source simulation environment called Google Research Football (GRF) \citep{kurach2020google}. GRF provides a standard platform to train and evaluate agents, and since its introduction, various methods have been developed to teach complex cooperative behaviors. However, it has been pointed out that a strategic gap exists between AI trained in simulations and the play of real-world players \citep{scott2021does}. To bridge this gap, more advanced learning methods have been proposed, such as \citep{lin2023tizero}, which successfully acquired human-level tactics by combining curriculum learning and self-play.

In the backward approach, which leverages real-world data, early research in ice hockey built Markov game models using play-by-play data to evaluate the contextual action values of players \citep{liu2018deep}. Similarly, in racket sports like badminton, deep reinforcement learning has been applied to player evaluation \citep{ding2022deep}. More recently, frameworks have been developed to support players' optimal decision-making from actual game data. For example, some studies have focused on optimizing a player's action selection to maximize the expected value of possession \citep{rahimian2022beyond}, while others have used inverse reinforcement learning to infer the underlying strategies of offensive and defensive plays \citep{rahimian2022inferring,luo2020inverse}. While these studies are effective for evaluating actions and strategies, \cite{nakahara2023action} is particularly notable for explicitly modeling individual players within a MARL framework, using actual tracking data to evaluate the action values of not only on-ball but also off-ball players.

While \cite{nakahara2023action} is a pioneering study that applied MARL to real tracking data to evaluate off-ball players, its state representation is based solely on the physical information (position and velocity) of the players and the ball. The novelty of our research lies in advancing this data-driven approach by proposing EDMS, an informationally and semantically dense state representation that explicitly incorporates tactically meaningful information. By doing so, we aim to improve upon conventional state representations that rely solely on physical information and to achieve more efficient policy learning and interpretable evaluation values.

\section{Deep Reinforcement Learning with Expandable Decision-Making States and Action Masking}
\label{sec:method}
In this section, we first describe our problem setting for our RL framework.
Then we propose our EDMS, describe action and reward definition, and propose our DRL with action masking.

\subsection{Problem Setting}
\label{ssec:problem_setting}
This study introduces an evaluation approach based on RL that calculates the value of player actions. As the final evaluation of plays, we use the state-action value (Q-value), which indicates the value of an action in a specific situation (i.e., state) according to the previous work \citep{nakahara2023action}.

To build the RL model, we first formulate the soccer play process as a Markov Decision Process (MDP). An MDP is a mathematical framework defined by a state, action, reward, and state transition probability. In this study, we apply this to the context of soccer and model it as a series of processes where, in a given situation (state \textit{s}), a player selects a specific play (action \textit{a}), and as a result, an evaluation value (reward \textit{r}) is given to the team. Here, we assume the Markov property, which states that the next state depends only on the current state and action, and not on the past history. One unit of learning (an episode) is defined as the period from the start to the end of a single ball possession. This allows the continuous flow of play to be handled naturally. Since possessions vary in duration, episodes have variable lengths.

The model in this paper extends this MDP framework as an independent multi-agent reinforcement learning (I-MARL), where each player makes decisions independently. However, since team performance is important, the reward is set to be shared by the entire team, and this encourages implicit cooperative behavior. Under this premise, the following discussion will proceed by omitting the subscripts that denote individual players.

Based on the previous work \citep{nakahara2023action}, the learning objective of the model in the MDP is to discover a policy ($\pi$) that maximizes the sum of rewards obtained within a single episode. Here, in order to treat all plays within a possession as equivalent without discounting them over time, the discount factor $\gamma$ is set to $1$.

\subsection{Expandable Decision-Making States}
\label{ssec:edms}
We explain how we defined the state in our RL model in an expandable and interpretable format called EDMS.
In soccer, the attacking team aims for the goal by finding open space and developing a quick attack before the defensive players return. On the other hand, the defensive team prevents attacks by filling the space held by the opponent and reducing their options, without giving them time to make decisions. In essence, soccer is a sport of competing for time and space. 
``Expandable'' means that EDMS is built from modular, per-player decision-making descriptors rather than a single vector of coordinates. This modular design lets practitioners extend the state to accommodate more players, richer tactical cues, or provider-specific variables while preserving a consistent per-agent schema and keeping downstream RL architectures unchanged.
We determined the EDMS that express this principle based on feedback from the coach of a Japanese soccer team.

\textbf{Overall structure.}
First, we explain the overall structure of EDMS, which has the following structure (also shown in Table \ref{tab:edms_details}).
First, the state is divided into \textit{Absolute State}, which consists of state variables that are static for each player, and \textit{Relative State}, which changes relative to each player. Next, Relative State is divided into \textit{On-ball State} for the player in possession of the ball, and \textit{Off-ball State} for players not in possession of the ball. Furthermore, On-ball State is divided into \textit{Intra-Possession State} for situations where possession is maintained within the same team, and \textit{Inter-Possession State} for situations where possession transfers between teams. In the Intra-Possession State, for a sequence of two actions, the player who performs the first action is considered the on-ball player, and the other players are considered off-ball players.

\begin{table*}[h!]
\centering
\caption{This table outlines the features used in the EDMS. The features are categorized into Relative State, which describes player relationships, and Absolute State, which provides the game context. Relative states are further subdivided into On-ball, Off-ball, Intra-Possession, and Inter-Possession states to capture nuanced tactical situations.}
\label{tab:edms_details}
\begin{tabular}{p{0.21\textwidth} p{0.21\textwidth} p{0.21\textwidth} p{0.21\textwidth}}
\toprule
\multicolumn{3}{c}{\textbf{Relative State}} & \textbf{Absolute State} \\
\cmidrule(r){1-3}
\multicolumn{2}{c}{On-ball State} & Off-ball State & \\
\cmidrule(r){1-2} \cmidrule(l){3-3}
\raggedright Intra-Possession State & \raggedright Inter-Possession State & & \\
\midrule
\raggedright The time for the nearest opponent to reach the ball & \raggedright The time for the closest player from each team to reach the ball & \raggedright The distance between the off-ball player and the ball & \raggedright The distance from the ball to each team's offside line \\
\addlinespace
\raggedright The distance to the opponent’s goal & \raggedright The distance from the ball to each team’s goal & \raggedright The time for the nearest opponent to reach the off-ball player & \raggedright Formation \\
\addlinespace
\raggedright The angle to the opponent’s goal & \raggedright The angle from the ball to each team’s goal & \raggedright The time for the nearest opponent to reach the pass lane to the off-ball player & \\
\addlinespace
Dribble score & Ball Speed & Space score & \\
\addlinespace
Shot score & Transition & \raggedright The change in the space score when moving 1 meter in 8 directions & \\
\addlinespace
Long ball score & & & \\
\bottomrule
\end{tabular}
\end{table*}

\textbf{Absolute State.}
Next, we explain the details of each state.
Absolute State includes the distance from the ball to each team's offside line and the formation. The offside line is the halfway line if all attacking players are in their own half; otherwise, it is the line of the second-to-last opponent relative to the defending team and the ball. The formation uses the information from the data as is; therefore, it does not account for changes in formation between attack and defense or formation changes due to substitutions.

\textbf{Relative State (Off-ball State).}
For the Relative State, first, we explain the Off-ball State, which consists of a total of six variables. Each state variable is calculated for each of the 10 off-ball players on the attacking team. 
The first is the distance between the off-ball player and the ball ($dist\_ball$). 
The second is the time for the nearest opponent to reach the off-ball player ($time\_to\_reach\_player$). 
The third is the time for the nearest opponent to reach the pass lane to the off-ball player ($time\_to\_reach\_passline$). 

\begin{figure}[!h]
    \centering
    \includegraphics[scale=0.55]{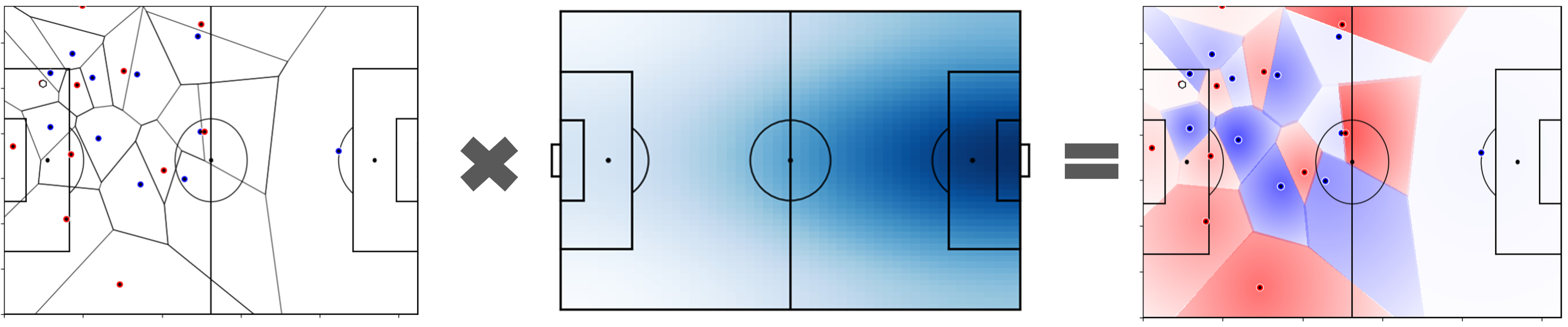}
    \caption{This figure illustrates the procedure for calculating the space score. From left to right, it shows: the Voronoi diagram (spatial area) considering each player's position and velocity; the importance of each area on the pitch; and the final space score distribution, which is the product of the two. The area importance is modeled based on proximity to the opponent's goal (sigmoid function) and distance from the center of the pitch (Gaussian function), and it is maximized in the central area in front of the opponent's goal}
    \label{fig:space_score}
\end{figure}

The fourth is the $space\_score$ (Fig. \ref{fig:space_score}). This is a value obtained by multiplying the weight of a space, which is a Voronoi region considering each player's position and velocity, by the importance of that area on the soccer field. The importance of the field is modeled based on two principles of attack. First, because the probability of scoring increases closer to the opponent's goal, the vertical value towards the goal is represented by a sigmoid function. This causes the value to be low near one's own goal and to increase sharply in an S-shape as it approaches the opponent's goal. Second, because the center of the field offers more play options and favorable shooting angles, the horizontal value based on the distance from the center is represented by a Gaussian function. This causes the value to be maximal at the center and to decrease smoothly towards the sidelines. The final importance is calculated by multiplying these vertical and horizontal values at each point, and its distribution is maximized in the central area in front of the opponent's goal. Since this score can also be calculated for the defensive team, it is possible to model which defensive players are dangerous when the attacking team loses the ball.

The fifth is the change in the space score when moving $1$ meter in $8$ directions. Here, the direction towards the opponent's goal ($x$-axis) is $0$ degrees, and the directions are divided in $45$-degree increments. 
The sixth is the pass\_score. The pass score is calculated from other off-ball state variables as follows.

\begin{align}
f_{pass\_score} = {}& k_1 dist\_ball + k_2 space\_score \notag \\
                    & + k_3 time\_to\_reach\_player \notag \\
                    & + k_4 time\_to\_reach\_passline.
\label{eq:pass_score}
\end{align}

Each weight $k$ was determined by a preliminary experiment that predicted pass success. In the preliminary study \citep{ide2024interpretable}, the success or failure of passes for the three teammates closest to the ball using the four variables required for the pass score were predicted and the importance of each variable was confirmed. According to the preliminary results, $k_1$ was set to $0.5$, $k_2$ to $0.3$, $k_3$ to $0.2$, and $k_4$ to $0.2$.

\textbf{Relative State (On-ball State).}
Next, we explain the On-ball State, which consists of Intra-Possession and Inter-Possession State.

The Intra-Possession State consists of a total of six state variables. 
The first is the time for the nearest opponent to reach the ball. 
The second is the distance to the opponent's goal. 
The third is the angle to the opponent's goal. 
The fourth is the $dribble\_score$, which is the change in the space score if the on-ball player moves 1 meter in 8 directions. 
The fifth is the $shot\_score$. The shot score is calculated by following these steps. First, only defending players, excluding the goalkeeper, who are located within the triangular area formed by connecting the on-ball player and both goalposts are considered. Next, we model the probability of each defender blocking a specific angle on the shot path. Finally, we calculate the overall block probability by multiple defenders and determine the expected value of the block probability by numerically integrating over all angles of the goalmouth. This score is calculated only when within 30 meters of the opponent's goal. 
The sixth is the $long\_ball\_score$. This is a score where the field is divided into right, central, and left zones, and a value of 1 is assigned to the zone containing the tallest player.

The Inter-Possession State consists of a total of five state variables. 
The first is the time for the closest player from each team to reach the ball. 
The second is the distance from the ball to each team's goal. 
The third is the angle from the ball to each team's goal. 
The fourth is the ball speed. 
The fifth is transition. This is a variable that represents the change between offense and defense; a 1 is assigned to players transitioning from attack to defense, and a 0 is assigned to players transitioning from defense to attack.

These state variables are only for representing the fundamentals of soccer. Therefore, variables that should be considered in a real match may not be included. However, since EDMS is extensible, it is possible to add variables necessary for individual analyses, such as individual player passing abilities to the Intra-Possession State, pass maps between players to the Offball State, or the match score to the Absolute State.

\subsection{Action and Reward Definition}
\label{ssec:action-and-reward}
Here we first explain the definition of the action space. In invasive team sports like soccer, the actions of off-ball players, in particular, have a continuous action space. However, a high-dimensional, continuous action space makes it difficult for a reinforcement learning model to learn. Therefore, in this study, based on prior research \citep{nakahara2023action}, we defined the actions a player can take as a finite number of discrete actions.

The actions defined in this model consist of a total of $16$ types, comprising on-ball and off-ball actions. The specific breakdown is as follows.
As on-ball actions, which are actions the player in possession can execute, we define pass, through pass, shot, cross, dribble, and defensive action. The defensive action includes multiple defensive actions such as interception, clearance, tackle, and block. This is because our study focuses on evaluating offensive plays, so we integrated these defensive actions as a single action without subdividing them. As off-ball actions, which are actions a player not in possession can execute, we define moving in $8$ directions and staying. This design of an action space allows the agent to learn the next action to take from a clear set of choices.

Next, we explain the reward design.
Soccer is a low-scoring sport, with an average of only two to three goals per match. Consequently, a reward design that only considers the final outcomes of goals scored or conceded leads to a sparse reward problem, where the feedback necessary for learning is obtained very infrequently. To mitigate this issue, \cite{nakahara2023action} designed a reward function that, in addition to goals and concessions, incorporated Expected Possession Value (EPV)\footnote{https://github.com/Friends-of-Tracking-Data-FoTD/LaurieOnTracking} as a positional value representing situational advantage. EPV is a metric that calculates the expected probability of an attacking sequence culminating in a goal, based on the current coordinates of the ball. Since EPV is higher closer to the opponent's goal, it allows the model to incorporate the value of ending an attack in a more threatening position.

In their design, the reward was given only at the end of a possession sequence; intermediate rewards during the sequence were always $0$. The terminal reward was defined based on the outcome of the sequence: $+1$ for a goal, $-1$ for a concession in the subsequent sequence, and the EPV otherwise. This design made it possible to reward attacking plays for their contribution, even if they did not directly result in a goal. However, a limitation of this design was that if a shot occurred mid-sequence, the reward was $0$, preventing the probability and value of the potential goal from being reflected in the action-values of the preceding events. To address this, we extend the reward design of \cite{nakahara2023action} by also providing the EPV as a reward at the moment a shot is taken. This modification allows the model to properly propagate the probability and value of a goal back to the action-values of the events that led to the shot.

\subsection{Deep Reinforcement Learning with Action Masking}
Here we describe DRL with action masking.
We evaluate plays by calculating the state-action value. The general framework of this model follows that of prior research \citep{nakahara2023action}. First, we will describe the parts that are the same as in prior research. In reinforcement learning, optimal state-action value (Q-value) is calculated by solving the Bellman equation. Here, the Q-value is defined as the expected value of the sum of future discounted rewards obtained when, in a certain state, a certain action is taken, and the policy is followed thereafter. The update of the Q-function is performed using five variables: the current state, the selected action, the obtained reward, the next state, and the next action as follows.

\begin{align}
Q^*(s_t, a_t) = {}& E_{s_{t+1},r_{t+1}}[r_{t+1} + \gamma Q(s_{t+1},a_{t+1})|s_t,a_t] \notag \\
= {}& \sum_{r_{t+1}} P(r_{t+1}|s_t,a_t)r_{t+1} \notag \\
& + \gamma \sum_{s_{t+1}} P(s_{t+1}|s_t,a_t)Q(s_{t+1},a_{t+1}).
\label{eq:qvalue}
\end{align}

As the learning framework, SARSA (state-action-reward-state-action) is used based on the previous work \citep{nakahara2023action}, which is an on-policy temporal difference (TD) learning algorithm. Additionally, since soccer play is essentially time-series data, we employ a Recurrent Neural Network (RNN) to capture its temporal dependencies. The input to the model consists of a sequence of plays, and each frame is composed of state variables representing the state of players and the field, and a one-hot vector of the action performed by each player. For the RNN layer, we use a single-layer Gated Recurrent Unit (GRU), which is excellent for learning long-term dependencies, with a hidden state dimensionality of $64$. The preceding input layer is a $64$-dimensional fully connected layer with a ReLU activation function. In the output layer, we calculate the Q-values for each of the defined actions for a given input state $s$.
However, the framework of this prior study has a significant limitation: it does not distinguish between actions available to on-ball and off-ball players. In the model, all players, regardless of ball possession, select from the same action set. This permits unrealistic scenarios, such as an off-ball player far from the ball selecting ``shot'', which unnecessarily expands the search space and reduces learning efficiency.

This issue is particularly problematic for TD learning methods like SARSA. If the model assigns a high Q-value to an impossible action, it risks performing wasteful updates to penalize that action, which can degrade the quality of the final policy given limited training data.

To solve this problem, we introduce action masking to incorporate the domain knowledge that ``only on-ball players can perform on-ball actions''. Specifically, we apply a mask to the Q-value vector output by the model based on the player's state. For an on-ball player, a large negative value (e.g., $-9999$) is substituted for the Q-values of off-ball actions, and vice versa for an off-ball player. This process prevents the selection of impossible actions and is expected to improve the quality of the learned policy.

The training of this model is performed by minimizing a complex loss function that combines the following three components.
The first is the TD loss. The estimation of Q-values is done by minimizing the loss as follows. 

\begin{align}
    \label{eq:td-loss}
    L_{td} = \sum_{t\in T}(r_{t+1} + \gamma Q(s_{t+1}, a_{t+1}) - Q(s_t, a_t))^2.
\end{align}
This loss function learns such that the difference between the value derived from the current state $s_t$ and action $a_t$ and the sum of the immediate reward $r$ and the value derived from the next state $s_{t+1}$ and action $a_{t+1}$ becomes smaller. 

The second is the supervised action loss. It is a loss function that uses the actions of professional players from real data as training labels. This is a supplementary loss function introduced to efficiently learn Q-values from limited data. Specifically, it is obtained by calculating the cross-entropy error between the softmax probability distribution of the Q-values and the one-hot vector of the actual action as follows.

\begin{align}
    \label{eq:as-loss}
    L_{as} = -\sum_{t \in T}\boldsymbol{a_t} \log(\textrm{softmax}(\boldsymbol{Q}_{s_t}))
\end{align}

The third is an $L_1$ regularization term. We adopt it to suppress model overfitting since the dataset size is relatively small.
The loss function consists of the three aforementioned components as follows.

\begin{align}
L_{total} = L_{td} + \lambda_1L_{L1} + \lambda_2L_{as}.
\label{eq:total_loss}
\end{align}

Since $L_{td}$ aims to learn the ``optimal action'' to maximize the reward, the action actually taken by the player and the direction of optimization may not always align. On the other hand, $L_{as}$ causes the agent to have a tendency to maintain a specific action. This difference creates a trade-off, and the balance of which loss to prioritize becomes important in the model's learning process. Therefore, in this study, based on preliminary experiments \citep{yeung2025openstarlab}, we set $\lambda_1=0.001$ and $\lambda_2=0.05$. Additionally, we used the Adam optimizer with a learning rate of $0.001$.




\section{Experimental Results}
\label{sec:experiment}
In this section, we describe the experimental evaluation of our proposed state representation, EDMS. Our main analysis is conducted on a proprietary J.League dataset, where we first quantitatively validate the superiority of EDMS over the baseline PVS and then perform detailed qualitative analysis of team tactics and specific in-game actions.

Following this, we assess the model's generalizability and practical applicability using diverse datasets, including LaLiga, the FIFA World Cup, and SoccerNet. The versatility of EDMS is demonstrated through consistent quantitative performance on common metrics across these datasets, supplemented by qualitative case studies of specific game scenarios and a visualization of Q-values on broadcast footage.

\subsection{Dataset in Main Analysis} 
\label{sec:experimental-dataset}
For the main analysis, we used event and tracking data from a total of 45 matches in the 2023 Meiji Yasuda J1 League, provided by Data Stadium Inc. The event data contains information on discrete actions such as passes and shots, including the action type, player name, ID, jersey number, timestamp, on-field coordinates, and an ID for each possession sequence (referred to as ``attack history ID''). The tracking data includes the x and y coordinates of all players and the ball, player jersey numbers (with the ball designated as 0), and frame numbers.

In the tracking data, we calculated the velocity and acceleration of all players and the ball and imputed missing values. We then synchronized the event and tracking data based on their timestamps. Following a previous study \citep{nakahara2023action}, we normalized the direction of attack to be consistently to the right and segmented the data into attacking sequences. These sequences are defined based on the ``attack history ID'' from the event data. This ID is assigned to a series of plays from when a team gains possession (e.g., via a set piece or by winning the ball) until they lose it. The same ID is maintained even if the ball is immediately regained after a mistake, such as a misplaced pass. We also selected sequences with a minimum of 30 frames and a maximum of 600 frames.

For our experiments, we used 45 matches from the 2022 season as the training data, 5 matches from the 2023 season as the validation data, and the remaining 40 matches from 2023 as the test data for model and player/team evaluation.


\subsection{Results of Main Analysis}
\subsubsection{Validation of proposed state and model}
We describe the evaluation results for each model. 
As a baseline, we used the state variables that were used in the previous study \citep{nakahara2023action}. Since those state variables consisted solely of the positions and velocities of all players, we will hereafter refer to them as the PVS (Position and Velocity States). To investigate the effects of EDMS, the RL model and reward design are the same as those in EDMS.
In this study, we evaluated each state model based on two metrics. First, to measure the performance of the learned policy, we used the action loss, which corresponds to the cross-entropy loss in the task of classifying the $16$ types of actions. Second, to validate the convergence of the Q-values learned by the model, we used TD loss.
The models for comparison are the following four: the first is PVS without action masking, the second is PVS with action masking, the third is EDMS without action masking, the fourth is EDMS with action masking. Table \ref{tab:validation_results} summarizes the action loss and TD loss (described as $L_{as}$ and $L_{td}$ in the previous section) for each model.

\begin{table*}[h!]
\centering
\begin{tabular}{l c c}
\hline\hline 
Model & action loss & TD loss \\
\hline 
PVS w/o mask & $2.6170$ & $2.6664 \times 10^{-4}$ \\
PVS w/ mask & $2.1126$ & $2.7707 \times 10^{-4}$  \\
EDMS w/o mask & $2.4286$ & $3.4001 \times 10^{-4}$  \\
EDMS w/ mask & $2.1087$ & $2.3580 \times 10^{-4}$ \\
\hline\hline
\end{tabular}
\vspace{5pt}
\caption{Comparison of action loss and TD loss for each model. PVS is the baseline model, and EDMS is the proposed model. 'w/o mask' indicates the results without an action mask, while 'w/ mask' indicates the results with an action mask. For both metrics, a lower value signifies better model performance.}
\label{tab:validation_results}
\end{table*}

First, focusing on the presence or absence of an action mask, we can see that for both PVS and EDMS, the models with an action mask (w/ mask) show a significant decrease in action loss compared to the models without it (w/o mask). Next, comparing our proposed method, EDMS, with the baseline PVS, the best performance was achieved by EDMS with an action mask, which achieved the lowest values among the four models for both action loss ($2.1087$) and TD loss ($2.3580 \times 10^{-4}$).

These experimental results indicate the effectiveness of the action mask and the superiority of the proposed EDMS. First, applying the action mask reduced the action loss in both models (PVS: from $2.6170$ to $2.1126 $; EDMS: from $2.4286$ to $2.1087$). This suggests that by eliminating invalid action choices in advance, the action mask reduced the search space, enabling the agent to learn the optimal policy more efficiently.
Next, we focus on the superiority of EDMS. In the case without a mask, the action loss for EDMS ($2.4286$) was lower than that for PVS ($2.6170$), indicating that the proposed state variables are more effective for action selection. With a mask, the action loss was nearly equivalent for both models, though EDMS was slightly better. Furthermore, when using an action mask, the TD loss for EDMS was distinctly lower than for PVS. This result strongly suggests that EDMS, when combined with an action mask, becomes capable of accurately evaluating state values.

Here, we turn our attention to the TD loss in the ``w/o mask'' case, where PVS ($2.6664 \times 10^{-4}$) is lower than EDMS ($3.4001 \times 10^{-4}$). Although EDMS has fewer dimensions than the baseline PVS, it differs significantly in the quality of its information. Whereas PVS handles raw physical information like player and ball positions and velocities, EDMS is composed of variables designed to have tactical meaning, such as ``the time for the nearest opponent to reach the ball'' and ``space score''. This means EDMS is an informationally and semantically richer state representation compared to PVS. This characteristic is likely the reason for the increased TD loss under the no-mask condition. The high-quality tactical information provided by EDMS leads the agent to expect a high state value. However, without an action mask, the agent can select actions that are not permissible in a given situation, creating a large discrepancy, a TD error, between the expected and actual values. On the other hand, by applying an action mask and appropriately constraining actions, this high-quality information can be used effectively, dramatically improving the accuracy of value estimation. In fact, EDMS with a mask recorded the lowest TD loss among the four models, indicating that the combination of our proposed state variables and an action mask is highly effective for enhancing the reinforcement learning agent's situational judgment capabilities.

\subsubsection{Analysis of Team Tactical Characteristics}
\begin{figure}[!h]
    \centering
    \includegraphics[scale=0.48]{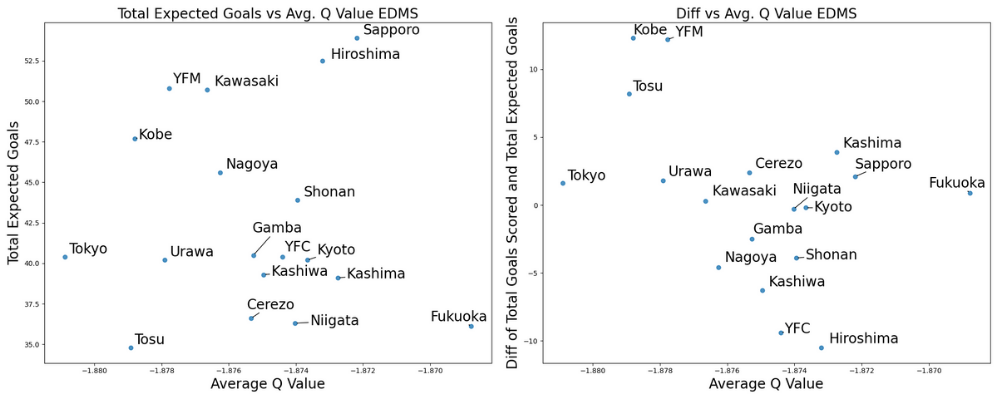}
    \caption{Relationship between each team's Q-value and various attacking metrics. The left plot shows the relationship with the total season Expected Goal (xG), and the right plot shows the relationship with the difference (Diff) between total season goals and xG.}
    \label{fig:stats_qvalue}
\end{figure}

To validate the effectiveness of the Q-values calculated using the proposed state representation EDMS and an action mask, we analyzed their relationship with existing soccer statistical indicators (Fig \ref{fig:stats_qvalue}). For comparison, we used Expected Goals (xG) and ``Diff'', the difference between actual goals and xG. The xG and Diff data were referenced from the entire season.
These data were referenced from the official J.League document ``J STATS REPORT 2023''\footnote{\url{https://jlib.j-league.or.jp/-site_media/media/content/82/2/index.html}}.

The analysis revealed no significant correlation between the Q-values and the existing stats ($p>0.05$, $r<=0.19$). This result suggests that the Q-values calculated in this study may be learning the non-linear value of plays that traditional stats fail to capture. Whereas xG and goals primarily evaluate the final actions, such as the shooting phase, the Q-value is thought to comprehensively evaluate the entire attacking process leading up to that point.

Next, we will conduct a detailed analysis of three teams (Consadole Sapporo, Sanfrecce Hiroshima, and Avispa Fukuoka) whose characteristic tactical styles emerged from the relationship between their Q-values and the existing stats.

Consadole Sapporo ranked among the highest in the league for both Q-value and xG. This indicates that their high-value attacking processes are directly linked to the creation of high-quality shooting opportunities. Furthermore, with a Diff value near zero, they are a team that executes an ideal offense, steadily converting the chances they create into goals as expected. According to ``Football Lab'' \footnote{https://www.football-lab.jp/} a data site provided by Data Stadium Inc., Sapporo's indices for short and long counters are among the best in the league, suggesting that our Q-value model may assign highly value to fast-paced counter-attacking tactics.

Sanfrecce Hiroshima also has high levels of both Q-value and xG, indicating they, like Sapporo, are capable of building high-quality attacking processes. However, they are characterized by a remarkably low Diff value within the league. This shows that while their tactics are functional and consistently create high-quality shooting chances, they face a serious problem with finishing in the final phase, causing them to fall significantly short of their expected goal tally. It is a classic case of ``a good process that doesn't yield results''.

Avispa Fukuoka showed a peculiar result: despite having the highest Q-value in the league, their xG was in the lower ranks. This signifies a disconnect between their process and the final phase, meaning that while they are the best at creating valuable situations likely to lead to a goal, these situations are not translating into high-quality shooting opportunities. Possible causes could be issues with the precision of the play just before the shot (the final pass or cross), or situations where they penetrate the penalty area but lose the ball before a shot can be taken. On the other hand, with a Diff value near zero, they also possess the efficiency to reliably convert their few chances into goals. This suggests they are a team with the potential to dramatically increase their scoring ability if they can improve the quality of their final pass.

\subsubsection{Evaluation of the Validity of Soccer Players' Off-the-ball Behavior Using the Q-values}
\begin{figure}[!h]
    \centering
    \includegraphics[scale=0.6]{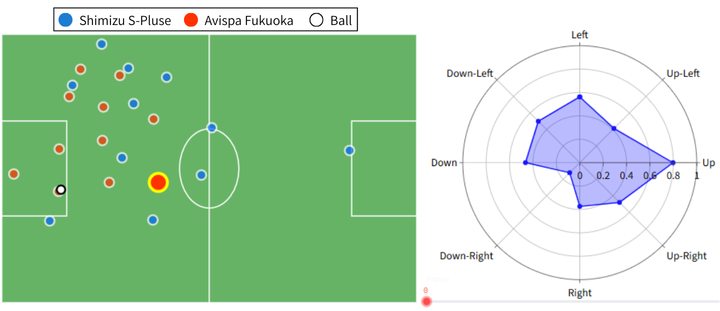}
    \caption{A build-up scene from the match between Shimizu S-Pulse and Avispa Fukuoka. The left plot shows the positions of all players and the ball, with Avispa Fukuoka attacking from left to right. The right plot shows the respective Q-values for the eight directional movements of the players enclosed in the yellow box on the left.}
    \label{fig:buildup_qvalue}
\end{figure}

Next, we qualitatively verified whether the calculated Q-values could evaluate tactically sound actions in specific in-game situations. In this analysis, as shown in Figure \ref{fig:buildup_qvalue}, we visualized a field plot on the left showing the position of all players and the ball, and on the right, the Q-values for the eight directional movements of a specific off-ball player, indicated by the yellow box in the left figure. Note that in all figures, the direction of attack is standardized to the right.

For our analysis, we focus on a situation involving a forward (FW) from Avispa Fukuoka during the build-up phase. Looking at the Q-value distribution in the right figure, the ``Up'' and ``Up-Right'' directions show high values. This represents movement towards the goal by utilizing the open space ahead, suggesting it is a rational choice for maximizing the reward of scoring.

On the other hand, the Q-values for the ``Left'', ``Down-Left'' and ``Down'' directions also show relatively high values. These movements can be interpreted as actions to reduce the distance to the ball-holder and other teammates, thereby providing support. By closing the distance between players, these movements are expected to secure passing lanes to facilitate combination play and to secure advantageous positions for immediate ball recovery in case of possession loss.

\begin{figure}[!h]
    \centering
    \includegraphics[scale=0.6]{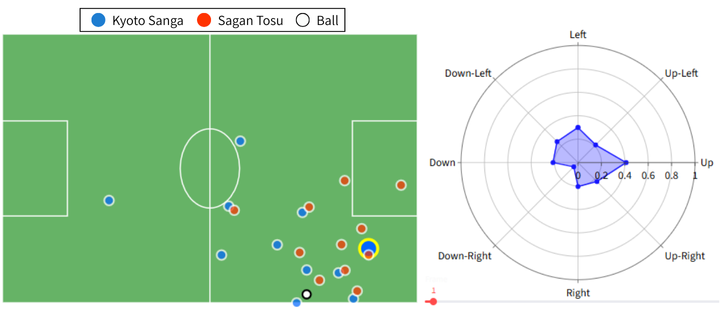}
    \caption{An offside situation from the match between Kyoto Sanga and Sagan Tosu, showing the moment before the pass was made. The figure layout is the same as in Figure \ref{fig:buildup_qvalue}}
    \label{fig:offside1_qvalue}
\end{figure}

\begin{figure}[!h]
    \centering
    \includegraphics[scale=0.6]{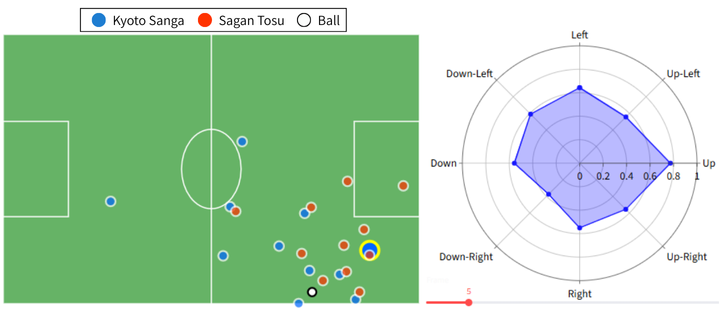}
    \caption{The same offside situation as in Figure \ref{fig:offside1_qvalue}, showing the moment after the pass was made. The figure layout is the same as in Figure \ref{fig:buildup_qvalue}.}
    \label{fig:offside2_qvalue}
\end{figure}

Next, we analyze the Q-values of a FW for Kyoto Sanga F.C. during an attacking phase in the opponent's territory. Figures \ref{fig:offside1_qvalue} and \ref{fig:offside2_qvalue} capture different moments (frames 2 and 5) within the same sequence.

In Figure \ref{fig:offside1_qvalue}, which captures the moment before a pass, the FW is in an offside position, resulting in low overall Q-values. In contrast, Figure \ref{fig:offside2_qvalue} shows the moment immediately after the pass is made. Now considered onside, the FW's Q-values have clearly increased.This dynamic change is explained by our EDMS state representation. Specifically, the space\_score variable is calculated as $0$ for any player in an offside position, correctly lowering their perceived Q-values. Once the ball is played, the FW becomes a valid onside target, their space\_score takes a positive value, and their Q-values are re-evaluated upwards. This demonstrates that our model successfully incorporates the offside rule to dynamically assess a player's value based on the context of the play.

\begin{figure}[!h]
    \centering
    \includegraphics[scale=0.6]{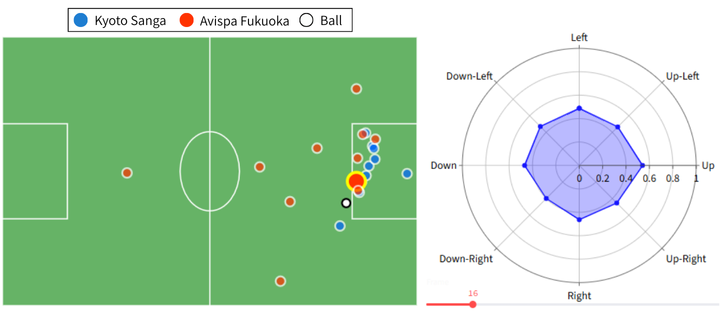}
    \caption{A Freekick scene from the match between Vissel Kobe and Avispa Fukuoka. The figure layout is the same as in Figure \ref{fig:buildup_qvalue}.}
    \label{fig:freekick_qvalue}
\end{figure}

Finally, we analyze the Q-values during a free kick situation for Avispa Fukuoka (Figure \ref{fig:freekick_qvalue}). In this scenario, the Q-values for the eight directional movements of the off-ball player were nearly uniform, resulting in the model being unable to suggest a clear optimal action.

This result suggests a limitation of the proposed method in special situations like set pieces. In scenarios such as free kicks or corner kicks, where players from both teams are crowded into a specific area and the play is static, the key state variables that constitute EDMS, such as space\_score and dist\_ball, are unlikely to produce significant differences among the action choices (eight directions of movement). We theorize that because there was no significant variation in the input state variables, the output Q-values were also similar, making it difficult for the model to evaluate the relative merits of each action.

This is a future challenge for our model, and we believe that improvements, such as adding state variables specifically designed for set pieces, will enable more robust evaluations across a wider variety of situations.

\subsection{Multi-Dataset Applicabiity through Modular Library Integration}
In this study, we integrated our defined state variables, EDMS, and a new dataset processing feature into the OpenSTARLab RLearn Package \citep{yeung2025openstarlab}, a deep learning library. In this section, we describe the results and applicability of this integration.

\subsubsection{Datasets}
This study utilized four distinct datasets for training and evaluation, comprising proprietary, commercial, and open-source data to ensure an analysis. The primary training dataset consists of $45$ matches from the $2023$ J1 League season, provided by DataStadium Inc., as detailed in Section 4.1. In addition, we incorporated commercial datasets for the LaLiga season, using data from StatsBomb and SkillCorner.
To enhance the reproducibility and test the model's versatility, two publicly available open-source datasets were also integrated: the FIFA World Cup 2022 dataset and SoccerNet GSR \citep{somers2024soccernet}. This allows for training and evaluation using a diverse range of match data.

The FIFA World Cup 2022 dataset was used for both training and evaluation. It contains event data and tracking data for all players from $64$ matches at a frame rate of $30$ fps. The coordinate system's origin is at the center of the pitch, with the x-axis representing the horizontal direction and the y-axis pointing towards the goal line. The pitch size is a standard 105m×68m.

On the other hand, the SoccerNet GSR dataset, which contains broadcast video from $164$ matches, was used exclusively as a test set for visualization purposes. Because broadcast video does not consistently capture all players on the field, complete tracking data is not available, making it unsuitable for training our model. Therefore, this dataset was used solely to overlay the Q-values calculated by our trained model onto actual video footage.

\subsubsection{OpenSTARLab RLearn Package}
The OpenSTARLab RLearn package \citep{yeung2025openstarlab} is a reinforcement learning (RL) toolkit designed for the time-step analysis of multi-agent decision-making models in soccer. The package specifically focuses on providing a comprehensive RL solution for soccer by integrally handling both event and tracking data.

This package uses PVS, the baseline in our study, as its state variables. The actions and evaluation metrics are also the same. On the other hand, the package supports the loading and preprocessing of the State-action-Reward (SAR) format, which integrates event and tracking data for RL tasks. This preprocessing is performed by the OpenSTARLab-Preprocessing package. Furthermore, our study employed a GRU, following the previous work \citep{nakahara2023action}. However, the RLearn package also has implementations of MLP (Multi-Layer Perceptron) and LSTM (Long Short-Term Memory), enabling comparative analysis with different architectures.

\subsubsection{Quantitative Validation for Each Dataset}
We verified whether the newly added open-source dataset (FIFA World Cup 2022) functions correctly within the OpenSTARLab framework, similar to the existing proprietary datasets (J1 League, LaLiga). Using each dataset, we trained both the baseline PVS and the proposed EDMS models and compared their results.

Table \ref{tab:openstarlab_results} shows the action loss and TD loss for PVS and EDMS trained on the three different datasets. The results confirm that the proposed method, EDMS, consistently outperformed the baseline PVS across all datasets. In both action loss and TD loss metrics, EDMS recorded lower values than PVS.

\begin{table*}[h!]
\centering
\begin{tabular}{l c c}
\hline\hline 
Data & action loss & TD loss \\
\hline 
J1 League PVS & $2.1126$ & $2.7707 \times 10^{-4}$  \\
J1 League EDMS & $2.1087$ & $2.3580 \times 10^{-4}$ \\
LaLiga PVS & $2.1468$ & $4.2908 \times 10^{-4}$ \\
LaLiga EDMS & $2.1344$ & $3.3545 \times 10^{-4}$ \\
FIFA World Cup PVS & $2.3016$ & $4.0331 \times 10^{-4}$ \\
FIFA World Cup EDMS & $2.2236$ & $3.0554 \times 10^{-4}$ \\
\hline\hline
\end{tabular}
\vspace{5pt}
\caption{Comparison of action loss and TD loss for each dataset. PVS is the baseline model, and EDMS is the proposed model.}
\label{tab:openstarlab_results}
\end{table*}

This outcome suggests two important points. First, the proposed method EDMS has high generalizability. EDMS demonstrated consistently superior performance over PVS on data with different characteristics-league matches (J1 League, LaLiga) and a short-term tournament (FIFA World Cup). This indicates that EDMS captures universal tactical values in soccer that are not dependent on a specific style of play.

Second, the integration of the open-source dataset was successful. The fact that the newly added dataset produced learning results with trends similar to those of the existing datasets demonstrates that the preprocessing by the OpenSTARLab-Preprocessing package is functioning correctly and that diverse data sources can be handled in a unified manner. This significantly enhances the reproducibility and scalability of future research.

\subsubsection{Qualitative Assessment of Off-Ball Behavior}
Similar to Section 4.2.3, we conducted an analysis of off-ball actions using Q-values on the La Liga and World Cup datasets.

First, we analyze an attacking scene for Real Madrid in a La Liga match against FC Barcelona. In this situation, there is space in front of the FW, as well as to the rear left and right. Looking at the Q-values, three directions show particularly high values: ``top-right'', ``bottom-right'', and ``bottom-left''.
The movement in the ``top-right'' direction is a run into the forward space; this is a rational choice to maximize expected goals, as a pass here could lead to a decisive chance. The movement in the ``bottom-left'' direction can be interpreted as a supportive move to get closer to the ball-holder and become a safer passing option. This evaluates the value of securing possession and resetting the attack. The movement in the ``bottom-right'' direction showed a high Q-value despite a teammate already being present in that area. While it is difficult to specify the exact reason for this, it is possible that the model is evaluating a complex value, such as confusing the opponent's defense by having multiple teammates run into the same area or preparing for a second ball.

\begin{figure}[!h]
    \centering
    \includegraphics[scale=0.6]{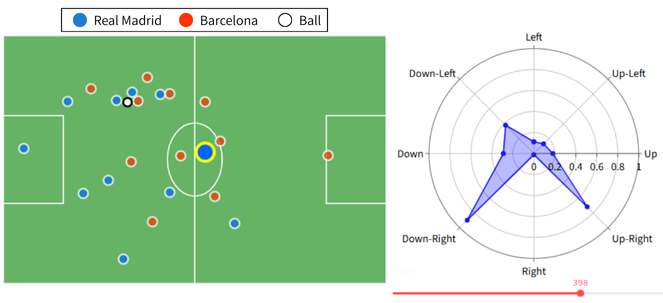}
    \caption{A attacking scene from the match between Real Madrid and Barcelona. The figure layout is the same as in Figure \ref{fig:buildup_qvalue}.}
    \label{fig:laliga_qvalues}
\end{figure}

Next, we analyze an attacking situation for Argentina near the touchline in the FIFA World Cup match against Australia. In this case, movements in the ``up'', ``top-left'', and ``left'' directions show high Q-values. All of these are movements away from the opposing defenders to secure space to receive a pass. Because the space in these directions is large, they have a high potential to become scoring opportunities. Additionally, the movement in the ``down'' direction also shows a relatively high value. This can be interpreted as a setup for a combination paly, such as moving toward the ball-holder to create a passing lane and draw a defender, then moving again into a different space. These results suggest that this model can evaluate not only movements that lead directly to a goal but also off-ball movements that create chances as tactically valuable plays.

\begin{figure}[!h]
    \centering
    \includegraphics[scale=0.55]{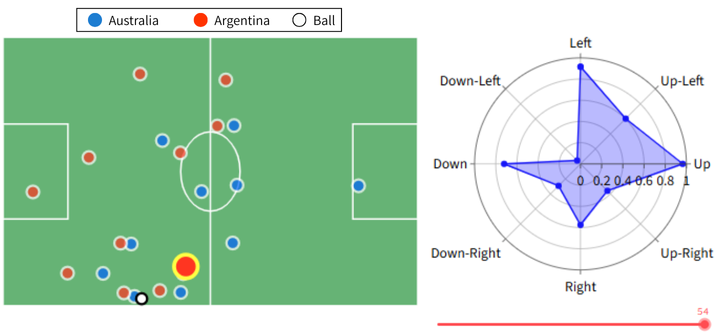}
    \caption{A defensive transition scene from the match between Australia and Argentina. The figure layout is the same as in Figure \ref{fig:buildup_qvalue}.}
    \label{fig:fifawc_qvalues}
\end{figure}

\subsubsection{SoccerNet GSR Visualization}
\begin{figure}[!h]
    \centering
    \includegraphics[scale=0.18]{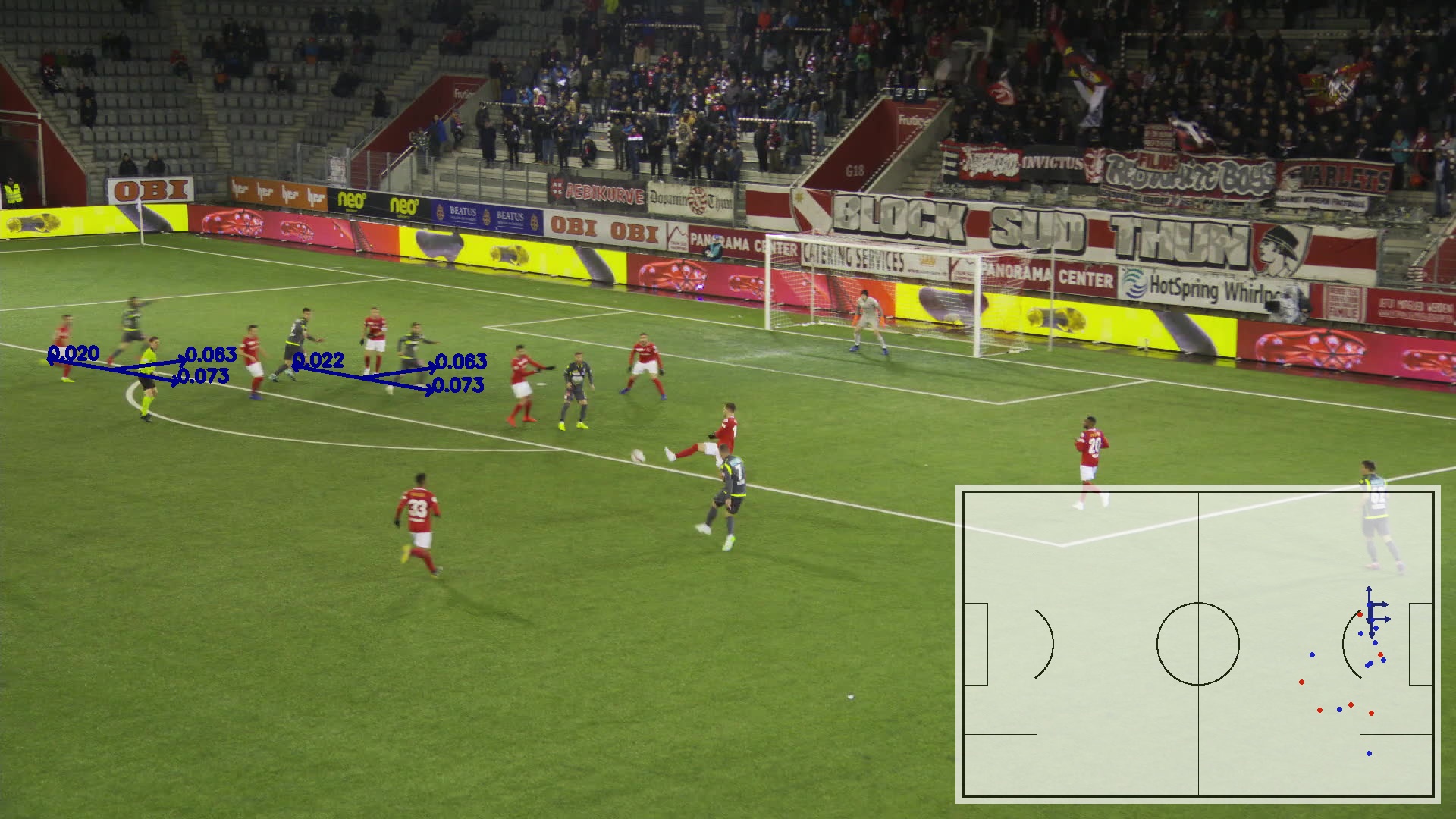}
    \caption{An example of projecting Q-values onto SoccerNet broadcast footage, showing a crossing scene. For the two players indicated, the arrows visualize the top three Q-values for their eight directional movements.}
    \label{fig:gsr}
\end{figure}

To demonstrate the practical applicability of the Q-values calculated in this study, we conducted a validation test by projecting them onto actual match footage using the SoccerNet broadcast video dataset. A challenge with broadcast footage is that it does not always capture all 22 players on the pitch, making it impossible to obtain the complete player position data required as input for the model. To address this issue, we applied a preprocessing step that assigned a default coordinate (0, 0), representing the center of the pitch, to players not visible in the frame.
Additionally, the event data was created manually. However, because manual annotation is labor-intensive, this validation used the FIFA World Cup data as the training and validation dataset, with only a single sequence from SoccerNet used as the test dataset.
Figure \ref{fig:gsr} is an example of the resulting projection of Q-values onto broadcast footage. In this figure, to improve visibility, we focused on two players and used arrows to indicate only the top three directions with the highest Q-values out of their eight possible movements. The analyzed scene is a moment when a cross is being delivered, and the Q-values for two players entering the penalty area are displayed. For both players, the movement toward the goal was calculated as having the maximum Q-value, suggesting that the model is correctly evaluating plays with a high probability of scoring.

Thus, the technology to project calculated Q-values onto actual broadcast footage holds the potential to contribute to enhanced fan engagement, not only as a tactical feedback tool for coaches and players but also by providing deeper match commentary for viewers.

\vspace{0pt}
\section{Conclusion}
\vspace{0pt}
\label{sec:conclusion}
In this study, we aimed to improve the performance of multi-agent deep reinforcement learning models in soccer by proposing and validating a new set of state variables, EDMS. Using the OpenSTARLab RLearn package as a foundation, we conducted comparative experiments against the baseline state variables, PVS, which primarily consist of physical information.

The experiments yielded the following three key findings. First, the action mask is extremely effective in policy learning, significantly reducing the action loss for both PVS and EDMS. Second, our proposed method, EDMS, when combined with an action mask, demonstrated the best performance, surpassing the baseline in both action loss and TD loss metrics.

Third, it was suggested that the superiority of EDMS stems from its high informational quality. Although EDMS has fewer dimensions than PVS, it is composed of informationally dense variables with tactical meaning, such as ``time for an opponent to reach the ball'' and ``space score''. While this characteristic caused a temporary increase in TD loss in an environment without an action mask, it conversely enabled the agent to evaluate the value of situations more accurately and learn the optimal policy more efficiently when actions were appropriately constrained. This indicates a strong synergistic effect between high-quality state representation and proper action constraints (i.e., the action mask).

The contribution of this research lies not only in proposing high-performance state variables but also in clarifying the effectiveness of its design philosophy—shifting from physical to semantic information—and the conditions required to maximize its utility. Furthermore, by integrating this method and newly added open-source datasets like the FIFA World Cup 2022 and SoccerNet into the OpenSTARLab framework, we contribute to the future development and improved reproducibility of soccer analytics research.

As for future prospects, the first is the application of more advanced multi-agent reinforcement learning models. While this study used independent RNNs, introducing models based on value decomposition methods like QMIX could potentially allow for more precise learning of the relationship between the team's overall value and individual players' values. This is expected to lead to a better evaluation of cooperative actions.

The second is the introduction of a game-theoretic approach. The interactions between players are based not just on simple cooperation or competition but on more complex relationships of interest. By using a game theory framework, it would become possible to model the strategic dependencies and decision-making among players, which would further deepen the evaluation of play.

\backmatter

\ifarxiv
\bmhead{Acknowledgments}
The J-league dataset used in this study was provided by the Research Organization of Information and Systems, The Institute of Statistical Mathematics and Data Stadium Inc.
We are grateful to the following supporters of our crowdfunding project (names given in no particular order): 
Eiji Konaka,
Junya Yatabe,
Yu Ukai,
Shotaro Ishihara,
Narita Youkan,
Takashi Hamaji,
Atom Scott,
Daichi Ishii,
Ikuma Uchida,
Yasuyuki Sawada,
Yosuke Yasuda,
Yumu Tabuchi,
Hiroyuki Shindo,
Kazuyuki Saka,
Red Dot Drone Japan,
Yuichiro Someya,
Keita Inoue,
Masato Max Yamamoto,
Rikuhei Umemoto,
Susumu Fukutome,
Takao Toshishige,
Makoto Isono,
Sho Yokoyama,
Leo Nakamura,
Masayuki Fukui, and
Hayato Ohtani.

\fi

\section*{Declarations}
\begin{itemize}
\item Funding: This work was financially supported by JST PRESTO Grant Number JPMJPR20CA, NEDO Grant Number 24021654, and JSPS KAKENHI Grant Number 23H03282.
\item Competing interests: The authors have no competing interests to declare that are relevant to the content of this article.
\item Ethics approval: Not applicable
\item Consent to participate: Not applicable
\item Consent for publication: Not applicable
\item Availability of data and materials: World Cup 2022 dataset and SoccerNet dataset are publicly available in \url{https://www.blog.fc.pff.com/blog/pff-fc-release-2022-world-cup-data} and \url{https://github.com/SoccerNet/sn-gamestate}. 
La Liga dataset is purchased, and J-league dataset is provided in a competition, so they cannot be shared. 
The code is avaiable at \url{https://github.com/open-starlab/}.
 
\ifarxiv
\item Authors' contributions: All authors contributed to the study conception and design. Data preparation, modeling, and analysis were performed by Kenjiro Ide. The first draft of the manuscript was written by Kenjiro Ide and Keisuke Fujii and all authors commented on previous versions of the manuscript. All authors read and approved the final manuscript.
\end{itemize}
\else
\fi

\begin{appendices}
\setcounter{table}{5}
\setcounter{figure}{6}
\setcounter{equation}{18}

\end{appendices}

\newpage
\bibliography{sn-article}

\begin{thebibliography}{}
\renewcommand{\doi}[1]{\url{https://doi.org/#1}}
\bibcommenthead

\bibitem [\protect \citeauthoryear {%
Brefeld%
, Lasek%
\BCBL {}\ \BBA {} Mair%
}{%
Brefeld%
\ \protect \BOthers {.}}{%
{\protect \APACyear {2019}}%
}]{%
brefeld2019probabilistic}
\APACinsertmetastar {%
brefeld2019probabilistic}%
\begin{APACrefauthors}%
Brefeld, U.%
, Lasek, J.%
\BCBL {} Mair, S.%
\end{APACrefauthors}%
\unskip\
\newblock
\APACrefYearMonthDay{2019}{}{}.
\newblock
{\BBOQ}\APACrefatitle {Probabilistic movement models and zones of control} {Probabilistic movement models and zones of control}.{\BBCQ}
\newblock
\APACjournalVolNumPages{Machine Learning}{108}{1}{127--147,}
\newblock

\newblock

\PrintBackRefs{\CurrentBib}

\bibitem [\protect \citeauthoryear {%
Decroos%
, Bransen%
, Van~Haaren%
\BCBL {}\ \BBA {} Davis%
}{%
Decroos%
\ \protect \BOthers {.}}{%
{\protect \APACyear {2019}}%
{\protect \APACexlab {{\protect \BCnt {1}}}}}]{%
Decroos19}
\APACinsertmetastar {%
Decroos19}%
\begin{APACrefauthors}%
Decroos, T.%
, Bransen, L.%
, Van~Haaren, J.%
\BCBL {} Davis, J.%
\end{APACrefauthors}%
\unskip\
\newblock
\APACrefYearMonthDay{2019{\protect \BCnt {1}}}{}{}.
\newblock
{\BBOQ}\APACrefatitle {Actions speak louder than goals: Valuing player actions in soccer} {Actions speak louder than goals: Valuing player actions in soccer}.{\BBCQ}
\newblock
 \APACrefbtitle {KDD} {Kdd}\ (\BPGS\ 1851--1861).
\PrintBackRefs{\CurrentBib}

\bibitem [\protect \citeauthoryear {%
Decroos%
, Bransen%
, Van~Haaren%
\BCBL {}\ \BBA {} Davis%
}{%
Decroos%
\ \protect \BOthers {.}}{%
{\protect \APACyear {2019}}%
{\protect \APACexlab {{\protect \BCnt {2}}}}}]{%
decroos2019actions}
\APACinsertmetastar {%
decroos2019actions}%
\begin{APACrefauthors}%
Decroos, T.%
, Bransen, L.%
, Van~Haaren, J.%
\BCBL {} Davis, J.%
\end{APACrefauthors}%
\unskip\
\newblock
\APACrefYearMonthDay{2019{\protect \BCnt {2}}}{}{}.
\newblock
{\BBOQ}\APACrefatitle {Actions speak louder than goals: Valuing player actions in soccer} {Actions speak louder than goals: Valuing player actions in soccer}.{\BBCQ}
\newblock
 \APACrefbtitle {Proceedings of the 25th ACM SIGKDD international conference on knowledge discovery \& data mining} {Proceedings of the 25th acm sigkdd international conference on knowledge discovery \& data mining}\ (\BPGS\ 1851--1861).
\PrintBackRefs{\CurrentBib}

\bibitem [\protect \citeauthoryear {%
Ding%
, Takeda%
\BCBL {}\ \BBA {} Fujii%
}{%
Ding%
\ \protect \BOthers {.}}{%
{\protect \APACyear {2022}}%
}]{%
ding2022deep}
\APACinsertmetastar {%
ding2022deep}%
\begin{APACrefauthors}%
Ding, N.%
, Takeda, K.%
\BCBL {} Fujii, K.%
\end{APACrefauthors}%
\unskip\
\newblock
\APACrefYearMonthDay{2022}{}{}.
\newblock
{\BBOQ}\APACrefatitle {Deep Reinforcement Learning in a Racket Sport for Player Evaluation With Technical and Tactical Contexts} {Deep reinforcement learning in a racket sport for player evaluation with technical and tactical contexts}.{\BBCQ}
\newblock
\APACjournalVolNumPages{IEEE Access}{10}{}{54764--54772,}
\newblock

\newblock

\PrintBackRefs{\CurrentBib}

\bibitem [\protect \citeauthoryear {%
Fernandez%
\ \BBA {} Bornn%
}{%
Fernandez%
\ \BBA {} Bornn%
}{%
{\protect \APACyear {2018}}%
}]{%
fernandez2018wide}
\APACinsertmetastar {%
fernandez2018wide}%
\begin{APACrefauthors}%
Fernandez, J.%
\BCBT {}\ \BBA {} Bornn, L.%
\end{APACrefauthors}%
\unskip\
\newblock
\APACrefYearMonthDay{2018}{}{}.
\newblock
{\BBOQ}\APACrefatitle {Wide Open Spaces: A statistical technique for measuring space creation in professional soccer} {Wide open spaces: A statistical technique for measuring space creation in professional soccer}.{\BBCQ}
\newblock
 \APACrefbtitle {Sloan sports analytics conference} {Sloan sports analytics conference}\ (\BVOL\ 2018).
\PrintBackRefs{\CurrentBib}

\bibitem [\protect \citeauthoryear {%
Fern{\'a}ndez%
, Bornn%
\BCBL {}\ \BBA {} Cervone%
}{%
Fern{\'a}ndez%
\ \protect \BOthers {.}}{%
{\protect \APACyear {2019}}%
}]{%
fernandez2019decomposing}
\APACinsertmetastar {%
fernandez2019decomposing}%
\begin{APACrefauthors}%
Fern{\'a}ndez, J.%
, Bornn, L.%
\BCBL {} Cervone, D.%
\end{APACrefauthors}%
\unskip\
\newblock
\APACrefYearMonthDay{2019}{}{}.
\newblock
{\BBOQ}\APACrefatitle {Decomposing the immeasurable sport: A deep learning expected possession value framework for soccer} {Decomposing the immeasurable sport: A deep learning expected possession value framework for soccer}.{\BBCQ}
\newblock
 \APACrefbtitle {13th MIT Sloan Sports Analytics Conference} {13th mit sloan sports analytics conference}\ (\BVOL~2).
\PrintBackRefs{\CurrentBib}

\bibitem [\protect \citeauthoryear {%
Fujii%
, Takeishi%
, Kawahara%
\BCBL {}\ \BBA {} Takeda%
}{%
Fujii%
, Takeishi%
\BCBL {}\ \protect \BOthers {.}}{%
{\protect \APACyear {2024}}%
}]{%
fujii2024decentralized}
\APACinsertmetastar {%
fujii2024decentralized}%
\begin{APACrefauthors}%
Fujii, K.%
, Takeishi, N.%
, Kawahara, Y.%
\BCBL {} Takeda, K.%
\end{APACrefauthors}%
\unskip\
\newblock
\APACrefYearMonthDay{2024}{}{}.
\newblock
{\BBOQ}\APACrefatitle {Decentralized policy learning with partial observation and mechanical constraints for multiperson modeling} {Decentralized policy learning with partial observation and mechanical constraints for multiperson modeling}.{\BBCQ}
\newblock
\APACjournalVolNumPages{Neural Networks}{171}{}{40--52,}
\newblock

\newblock

\PrintBackRefs{\CurrentBib}

\bibitem [\protect \citeauthoryear {%
Fujii%
, Takeuchi%
\BCBL {}\ \protect \BOthers {.}}{%
Fujii%
, Takeuchi%
\BCBL {}\ \protect \BOthers {.}}{%
{\protect \APACyear {2024}}%
}]{%
fujii2024estimating}
\APACinsertmetastar {%
fujii2024estimating}%
\begin{APACrefauthors}%
Fujii, K.%
, Takeuchi, K.%
, Kuribayashi, A.%
, Takeishi, N.%
, Kawahara, Y.%
\BCBL {} Takeda, K.%
\end{APACrefauthors}%
\unskip\
\newblock
\APACrefYearMonthDay{2024}{}{}.
\newblock
{\BBOQ}\APACrefatitle {Estimating counterfactual treatment outcomes over time in complex multiagent scenarios} {Estimating counterfactual treatment outcomes over time in complex multiagent scenarios}.{\BBCQ}
\newblock
\APACjournalVolNumPages{IEEE Transactions on Neural Networks and Learning Systems}{36}{2}{2103--2117,}
\newblock

\newblock

\PrintBackRefs{\CurrentBib}

\bibitem [\protect \citeauthoryear {%
Fujimura%
\ \BBA {} Sugihara%
}{%
Fujimura%
\ \BBA {} Sugihara%
}{%
{\protect \APACyear {2005}}%
}]{%
fujimura2005geometric}
\APACinsertmetastar {%
fujimura2005geometric}%
\begin{APACrefauthors}%
Fujimura, A.%
\BCBT {}\ \BBA {} Sugihara, K.%
\end{APACrefauthors}%
\unskip\
\newblock
\APACrefYearMonthDay{2005}{}{}.
\newblock
{\BBOQ}\APACrefatitle {Geometric analysis and quantitative evaluation of sport teamwork} {Geometric analysis and quantitative evaluation of sport teamwork}.{\BBCQ}
\newblock
\APACjournalVolNumPages{Systems and Computers in Japan}{36}{6}{49--58,}
\newblock

\newblock

\PrintBackRefs{\CurrentBib}

\bibitem [\protect \citeauthoryear {%
Ide%
, Someya%
, Kawaguchi%
\BCBL {}\ \BBA {} Fujii%
}{%
Ide%
\ \protect \BOthers {.}}{%
{\protect \APACyear {2025}}%
}]{%
ide2024interpretable}
\APACinsertmetastar {%
ide2024interpretable}%
\begin{APACrefauthors}%
Ide, K.%
, Someya, T.%
, Kawaguchi, K.%
\BCBL {} Fujii, K.%
\end{APACrefauthors}%
\unskip\
\newblock
\APACrefYearMonthDay{2025}{}{}.
\newblock
{\BBOQ}\APACrefatitle {Interpretable Low-Dimensional Modeling of Spatiotemporal Agent States for Decision Making in Football Tactics} {Interpretable low-dimensional modeling of spatiotemporal agent states for decision making in football tactics}.{\BBCQ}
\newblock
\APACjournalVolNumPages{arXiv preprint arXiv:2506.16696}{}{}{,}
\newblock

\newblock

\PrintBackRefs{\CurrentBib}

\bibitem [\protect \citeauthoryear {%
Kurach%
\ \protect \BOthers {.}}{%
Kurach%
\ \protect \BOthers {.}}{%
{\protect \APACyear {2020}}%
}]{%
kurach2020google}
\APACinsertmetastar {%
kurach2020google}%
\begin{APACrefauthors}%
Kurach, K.%
, Raichuk, A.%
, Sta{\'n}czyk, P.%
, Zaj{\k{a}}c, M.%
, Bachem, O.%
, Espeholt, L.%
\BDBL {}others%
\end{APACrefauthors}%
\unskip\
\newblock
\APACrefYearMonthDay{2020}{}{}.
\newblock
{\BBOQ}\APACrefatitle {Google research football: A novel reinforcement learning environment} {Google research football: A novel reinforcement learning environment}.{\BBCQ}
\newblock
 \APACrefbtitle {Proceedings of the AAAI Conference on Artificial Intelligence} {Proceedings of the aaai conference on artificial intelligence}\ (\BVOL~34, \BPGS\ 4501--4510).
\PrintBackRefs{\CurrentBib}

\bibitem [\protect \citeauthoryear {%
Le%
, Yue%
, Carr%
\BCBL {}\ \BBA {} Lucey%
}{%
Le%
\ \protect \BOthers {.}}{%
{\protect \APACyear {2017}}%
}]{%
le2017coordinated}
\APACinsertmetastar {%
le2017coordinated}%
\begin{APACrefauthors}%
Le, H.M.%
, Yue, Y.%
, Carr, P.%
\BCBL {} Lucey, P.%
\end{APACrefauthors}%
\unskip\
\newblock
\APACrefYearMonthDay{2017}{}{}.
\newblock
{\BBOQ}\APACrefatitle {Coordinated multi-agent imitation learning} {Coordinated multi-agent imitation learning}.{\BBCQ}
\newblock
 \APACrefbtitle {International Conference on Machine Learning} {International conference on machine learning}\ (\BPGS\ 1995--2003).
\PrintBackRefs{\CurrentBib}

\bibitem [\protect \citeauthoryear {%
Lin%
, Huang%
, Pearce%
, Chen%
\BCBL {}\ \BBA {} Tu%
}{%
Lin%
\ \protect \BOthers {.}}{%
{\protect \APACyear {2023}}%
}]{%
lin2023tizero}
\APACinsertmetastar {%
lin2023tizero}%
\begin{APACrefauthors}%
Lin, F.%
, Huang, S.%
, Pearce, T.%
, Chen, W.%
\BCBL {} Tu, W\BHBI W.%
\end{APACrefauthors}%
\unskip\
\newblock
\APACrefYearMonthDay{2023}{}{}.
\newblock
{\BBOQ}\APACrefatitle {Tizero: Mastering multi-agent football with curriculum learning and self-play} {Tizero: Mastering multi-agent football with curriculum learning and self-play}.{\BBCQ}
\newblock
\APACjournalVolNumPages{arXiv preprint arXiv:2302.07515}{}{}{,}
\newblock

\newblock

\PrintBackRefs{\CurrentBib}

\bibitem [\protect \citeauthoryear {%
Link%
, Lang%
\BCBL {}\ \BBA {} Seidenschwarz%
}{%
Link%
\ \protect \BOthers {.}}{%
{\protect \APACyear {2016}}%
}]{%
Link2016}
\APACinsertmetastar {%
Link2016}%
\begin{APACrefauthors}%
Link, D.%
, Lang, S.%
\BCBL {} Seidenschwarz, P.%
\end{APACrefauthors}%
\unskip\
\newblock
\APACrefYearMonthDay{2016}{}{}.
\newblock
{\BBOQ}\APACrefatitle {Real time quantification of dangerousity in football using spatiotemporal tracking data} {Real time quantification of dangerousity in football using spatiotemporal tracking data}.{\BBCQ}
\newblock
\APACjournalVolNumPages{PloS one}{11}{12}{e0168768,}
\newblock

\newblock

\PrintBackRefs{\CurrentBib}

\bibitem [\protect \citeauthoryear {%
Liu%
\ \BBA {} Schulte%
}{%
Liu%
\ \BBA {} Schulte%
}{%
{\protect \APACyear {2018}}%
}]{%
liu2018deep}
\APACinsertmetastar {%
liu2018deep}%
\begin{APACrefauthors}%
Liu, G.%
\BCBT {}\ \BBA {} Schulte, O.%
\end{APACrefauthors}%
\unskip\
\newblock
\APACrefYearMonthDay{2018}{}{}.
\newblock
{\BBOQ}\APACrefatitle {Deep reinforcement learning in ice hockey for context-aware player evaluation} {Deep reinforcement learning in ice hockey for context-aware player evaluation}.{\BBCQ}
\newblock
\APACjournalVolNumPages{arXiv preprint arXiv:1805.11088}{}{}{,}
\newblock

\newblock

\PrintBackRefs{\CurrentBib}

\bibitem [\protect \citeauthoryear {%
Lowe%
\ \protect \BOthers {.}}{%
Lowe%
\ \protect \BOthers {.}}{%
{\protect \APACyear {2017}}%
}]{%
lowe2017multi}
\APACinsertmetastar {%
lowe2017multi}%
\begin{APACrefauthors}%
Lowe, R.%
, Wu, Y.I.%
, Tamar, A.%
, Harb, J.%
, Pieter~Abbeel, O.%
\BCBL {} Mordatch, I.%
\end{APACrefauthors}%
\unskip\
\newblock
\APACrefYearMonthDay{2017}{}{}.
\newblock
{\BBOQ}\APACrefatitle {Multi-agent actor-critic for mixed cooperative-competitive environments} {Multi-agent actor-critic for mixed cooperative-competitive environments}.{\BBCQ}
\newblock
\APACjournalVolNumPages{Advances in neural information processing systems}{30}{}{,}
\newblock

\newblock

\PrintBackRefs{\CurrentBib}

\bibitem [\protect \citeauthoryear {%
Luo%
}{%
Luo%
}{%
{\protect \APACyear {2020}}%
}]{%
luo2020inverse}
\APACinsertmetastar {%
luo2020inverse}%
\begin{APACrefauthors}%
Luo, Y.%
\end{APACrefauthors}%
\unskip\
\newblock
\APACrefYearMonthDay{2020}{}{}.
\newblock
{\BBOQ}\APACrefatitle {Inverse reinforcement learning for team sports: Valuing actions and players} {Inverse reinforcement learning for team sports: Valuing actions and players}.{\BBCQ}
\newblock

\newblock

\newblock

\PrintBackRefs{\CurrentBib}

\bibitem [\protect \citeauthoryear {%
Nakahara%
, Tsutsui%
, Takeda%
\BCBL {}\ \BBA {} Fujii%
}{%
Nakahara%
\ \protect \BOthers {.}}{%
{\protect \APACyear {2023}}%
}]{%
nakahara2023action}
\APACinsertmetastar {%
nakahara2023action}%
\begin{APACrefauthors}%
Nakahara, H.%
, Tsutsui, K.%
, Takeda, K.%
\BCBL {} Fujii, K.%
\end{APACrefauthors}%
\unskip\
\newblock
\APACrefYearMonthDay{2023}{}{}.
\newblock
{\BBOQ}\APACrefatitle {Action valuation of on-and off-ball soccer players based on multi-agent deep reinforcement learning} {Action valuation of on-and off-ball soccer players based on multi-agent deep reinforcement learning}.{\BBCQ}
\newblock
\APACjournalVolNumPages{IEEE Access}{11}{}{131237--131244,}
\newblock

\newblock

\PrintBackRefs{\CurrentBib}

\bibitem [\protect \citeauthoryear {%
Ng%
, Russell%
\BCBL {}\ \protect \BOthers {.}}{%
Ng%
\ \protect \BOthers {.}}{%
{\protect \APACyear {2000}}%
}]{%
ng2000algorithms}
\APACinsertmetastar {%
ng2000algorithms}%
\begin{APACrefauthors}%
Ng, A.Y.%
, Russell, S.%
\BCBL {}\ \BOthersPeriod {.}\end{APACrefauthors}%
\unskip\
\newblock
\APACrefYearMonthDay{2000}{}{}.
\newblock
{\BBOQ}\APACrefatitle {Algorithms for inverse reinforcement learning.} {Algorithms for inverse reinforcement learning.}{\BBCQ}
\newblock
 \APACrefbtitle {Icml} {Icml}\ (\BVOL~1, \BPG~2).
\PrintBackRefs{\CurrentBib}

\bibitem [\protect \citeauthoryear {%
Nijssen%
\ \BBA {} De~Raedt%
}{%
Nijssen%
\ \BBA {} De~Raedt%
}{%
{\protect \APACyear {2006}}%
}]{%
nijssen2006iql}
\APACinsertmetastar {%
nijssen2006iql}%
\begin{APACrefauthors}%
Nijssen, S.%
\BCBT {}\ \BBA {} De~Raedt, L.%
\end{APACrefauthors}%
\unskip\
\newblock
\APACrefYearMonthDay{2006}{}{}.
\newblock
{\BBOQ}\APACrefatitle {Iql: A proposal for an inductive query language} {Iql: A proposal for an inductive query language}.{\BBCQ}
\newblock
 \APACrefbtitle {International Workshop on Knowledge Discovery in Inductive Databases} {International workshop on knowledge discovery in inductive databases}\ (\BPGS\ 189--207).
\PrintBackRefs{\CurrentBib}

\bibitem [\protect \citeauthoryear {%
Power%
, Ruiz%
, Wei%
\BCBL {}\ \BBA {} Lucey%
}{%
Power%
\ \protect \BOthers {.}}{%
{\protect \APACyear {2017}}%
}]{%
power2017not}
\APACinsertmetastar {%
power2017not}%
\begin{APACrefauthors}%
Power, P.%
, Ruiz, H.%
, Wei, X.%
\BCBL {} Lucey, P.%
\end{APACrefauthors}%
\unskip\
\newblock
\APACrefYearMonthDay{2017}{}{}.
\newblock
{\BBOQ}\APACrefatitle {Not all passes are created equal: Objectively measuring the risk and reward of passes in soccer from tracking data} {Not all passes are created equal: Objectively measuring the risk and reward of passes in soccer from tracking data}.{\BBCQ}
\newblock
 \APACrefbtitle {Proceedings of the 23rd ACM SIGKDD international conference on knowledge discovery and data mining} {Proceedings of the 23rd acm sigkdd international conference on knowledge discovery and data mining}\ (\BPGS\ 1605--1613).
\PrintBackRefs{\CurrentBib}

\bibitem [\protect \citeauthoryear {%
Rahimian%
\ \BBA {} Toka%
}{%
Rahimian%
\ \BBA {} Toka%
}{%
{\protect \APACyear {2022}}%
}]{%
rahimian2022inferring}
\APACinsertmetastar {%
rahimian2022inferring}%
\begin{APACrefauthors}%
Rahimian, P.%
\BCBT {}\ \BBA {} Toka, L.%
\end{APACrefauthors}%
\unskip\
\newblock
\APACrefYearMonthDay{2022}{}{}.
\newblock
{\BBOQ}\APACrefatitle {Inferring the strategy of offensive and defensive play in soccer with inverse reinforcement learning} {Inferring the strategy of offensive and defensive play in soccer with inverse reinforcement learning}.{\BBCQ}
\newblock
 \APACrefbtitle {Machine Learning and Data Mining for Sports Analytics: 8th International Workshop, MLSA 2021, Virtual Event, September 13, 2021, Revised Selected Papers} {Machine learning and data mining for sports analytics: 8th international workshop, mlsa 2021, virtual event, september 13, 2021, revised selected papers}\ (\BPGS\ 26--38).
\PrintBackRefs{\CurrentBib}

\bibitem [\protect \citeauthoryear {%
Rahimian%
, Van~Haaren%
, Abzhanova%
\BCBL {}\ \BBA {} Toka%
}{%
Rahimian%
\ \protect \BOthers {.}}{%
{\protect \APACyear {2022}}%
}]{%
rahimian2022beyond}
\APACinsertmetastar {%
rahimian2022beyond}%
\begin{APACrefauthors}%
Rahimian, P.%
, Van~Haaren, J.%
, Abzhanova, T.%
\BCBL {} Toka, L.%
\end{APACrefauthors}%
\unskip\
\newblock
\APACrefYearMonthDay{2022}{}{}.
\newblock
{\BBOQ}\APACrefatitle {Beyond action valuation: A deep reinforcement learning framework for optimizing player decisions in soccer} {Beyond action valuation: A deep reinforcement learning framework for optimizing player decisions in soccer}.{\BBCQ}
\newblock
 \APACrefbtitle {16th Annual MIT Sloan Sports Analytics Conference. Boston, MA, USA: MIT} {16th annual mit sloan sports analytics conference. boston, ma, usa: Mit}\ (\BPG~25).
\PrintBackRefs{\CurrentBib}

\bibitem [\protect \citeauthoryear {%
Rashid%
\ \protect \BOthers {.}}{%
Rashid%
\ \protect \BOthers {.}}{%
{\protect \APACyear {2020}}%
}]{%
rashid2020monotonic}
\APACinsertmetastar {%
rashid2020monotonic}%
\begin{APACrefauthors}%
Rashid, T.%
, Samvelyan, M.%
, De~Witt, C.S.%
, Farquhar, G.%
, Foerster, J.%
\BCBL {} Whiteson, S.%
\end{APACrefauthors}%
\unskip\
\newblock
\APACrefYearMonthDay{2020}{}{}.
\newblock
{\BBOQ}\APACrefatitle {Monotonic value function factorisation for deep multi-agent reinforcement learning} {Monotonic value function factorisation for deep multi-agent reinforcement learning}.{\BBCQ}
\newblock
\APACjournalVolNumPages{Journal of Machine Learning Research}{21}{178}{1--51,}
\newblock

\newblock

\PrintBackRefs{\CurrentBib}

\bibitem [\protect \citeauthoryear {%
Scott%
, Fujii%
\BCBL {}\ \BBA {} Onishi%
}{%
Scott%
\ \protect \BOthers {.}}{%
{\protect \APACyear {2021}}%
}]{%
scott2021does}
\APACinsertmetastar {%
scott2021does}%
\begin{APACrefauthors}%
Scott, A.%
, Fujii, K.%
\BCBL {} Onishi, M.%
\end{APACrefauthors}%
\unskip\
\newblock
\APACrefYearMonthDay{2021}{}{}.
\newblock
{\BBOQ}\APACrefatitle {How does AI play football? An analysis of RL and real-world football strategies} {How does ai play football? an analysis of rl and real-world football strategies}.{\BBCQ}
\newblock
\APACjournalVolNumPages{arXiv preprint arXiv:2111.12340}{}{}{,}
\newblock

\newblock

\PrintBackRefs{\CurrentBib}

\bibitem [\protect \citeauthoryear {%
Silver%
\ \protect \BOthers {.}}{%
Silver%
\ \protect \BOthers {.}}{%
{\protect \APACyear {2017}}%
}]{%
silver2017mastering}
\APACinsertmetastar {%
silver2017mastering}%
\begin{APACrefauthors}%
Silver, D.%
, Hubert, T.%
, Schrittwieser, J.%
, Antonoglou, I.%
, Lai, M.%
, Guez, A.%
\BDBL {}others%
\end{APACrefauthors}%
\unskip\
\newblock
\APACrefYearMonthDay{2017}{}{}.
\newblock
{\BBOQ}\APACrefatitle {Mastering chess and shogi by self-play with a general reinforcement learning algorithm} {Mastering chess and shogi by self-play with a general reinforcement learning algorithm}.{\BBCQ}
\newblock
\APACjournalVolNumPages{arXiv preprint arXiv:1712.01815}{}{}{,}
\newblock

\newblock

\PrintBackRefs{\CurrentBib}

\bibitem [\protect \citeauthoryear {%
Simpson%
, Beal%
, Locke%
\BCBL {}\ \BBA {} Norman%
}{%
Simpson%
\ \protect \BOthers {.}}{%
{\protect \APACyear {2022}}%
}]{%
simpson2022seq2event}
\APACinsertmetastar {%
simpson2022seq2event}%
\begin{APACrefauthors}%
Simpson, I.%
, Beal, R.J.%
, Locke, D.%
\BCBL {} Norman, T.J.%
\end{APACrefauthors}%
\unskip\
\newblock
\APACrefYearMonthDay{2022}{}{}.
\newblock
{\BBOQ}\APACrefatitle {Seq2event: Learning the language of soccer using transformer-based match event prediction} {Seq2event: Learning the language of soccer using transformer-based match event prediction}.{\BBCQ}
\newblock
 \APACrefbtitle {Proceedings of the 28th ACM SIGKDD conference on knowledge discovery and data mining} {Proceedings of the 28th acm sigkdd conference on knowledge discovery and data mining}\ (\BPGS\ 3898--3908).
\PrintBackRefs{\CurrentBib}

\bibitem [\protect \citeauthoryear {%
Singh%
}{%
Singh%
}{%
{\protect \APACyear {2018}}%
}]{%
singh2018introducing}
\APACinsertmetastar {%
singh2018introducing}%
\begin{APACrefauthors}%
Singh, K.%
\end{APACrefauthors}%
\unskip\
\newblock
\APACrefYearMonthDay{2018}{}{}.
\newblock
{\BBOQ}\APACrefatitle {Introducing Expected Threat (xT): Modelling team behaviour in possession to gain a deeper understanding of buildup play} {Introducing expected threat (xt): Modelling team behaviour in possession to gain a deeper understanding of buildup play}.{\BBCQ}
\newblock
\APACjournalVolNumPages{Consultado el}{7}{}{,}
\newblock

\newblock

\PrintBackRefs{\CurrentBib}

\bibitem [\protect \citeauthoryear {%
Somers%
\ \protect \BOthers {.}}{%
Somers%
\ \protect \BOthers {.}}{%
{\protect \APACyear {2024}}%
}]{%
somers2024soccernet}
\APACinsertmetastar {%
somers2024soccernet}%
\begin{APACrefauthors}%
Somers, V.%
, Joos, V.%
, Cioppa, A.%
, Giancola, S.%
, Ghasemzadeh, S.A.%
, Magera, F.%
\BDBL {}others%
\end{APACrefauthors}%
\unskip\
\newblock
\APACrefYearMonthDay{2024}{}{}.
\newblock
{\BBOQ}\APACrefatitle {SoccerNet game state reconstruction: End-to-end athlete tracking and identification on a minimap} {Soccernet game state reconstruction: End-to-end athlete tracking and identification on a minimap}.{\BBCQ}
\newblock
 \APACrefbtitle {Proceedings of the IEEE/CVF Conference on Computer Vision and Pattern Recognition} {Proceedings of the ieee/cvf conference on computer vision and pattern recognition}\ (\BPGS\ 3293--3305).
\PrintBackRefs{\CurrentBib}

\bibitem [\protect \citeauthoryear {%
Son%
, Kim%
, Kang%
, Hostallero%
\BCBL {}\ \BBA {} Yi%
}{%
Son%
\ \protect \BOthers {.}}{%
{\protect \APACyear {2019}}%
}]{%
son2019qtran}
\APACinsertmetastar {%
son2019qtran}%
\begin{APACrefauthors}%
Son, K.%
, Kim, D.%
, Kang, W.J.%
, Hostallero, D.E.%
\BCBL {} Yi, Y.%
\end{APACrefauthors}%
\unskip\
\newblock
\APACrefYearMonthDay{2019}{}{}.
\newblock
{\BBOQ}\APACrefatitle {Qtran: Learning to factorize with transformation for cooperative multi-agent reinforcement learning} {Qtran: Learning to factorize with transformation for cooperative multi-agent reinforcement learning}.{\BBCQ}
\newblock
 \APACrefbtitle {International conference on machine learning} {International conference on machine learning}\ (\BPGS\ 5887--5896).
\PrintBackRefs{\CurrentBib}

\bibitem [\protect \citeauthoryear {%
Sunehag%
\ \protect \BOthers {.}}{%
Sunehag%
\ \protect \BOthers {.}}{%
{\protect \APACyear {2017}}%
}]{%
sunehag2017value}
\APACinsertmetastar {%
sunehag2017value}%
\begin{APACrefauthors}%
Sunehag, P.%
, Lever, G.%
, Gruslys, A.%
, Czarnecki, W.M.%
, Zambaldi, V.%
, Jaderberg, M.%
\BDBL {}others%
\end{APACrefauthors}%
\unskip\
\newblock
\APACrefYearMonthDay{2017}{}{}.
\newblock
{\BBOQ}\APACrefatitle {Value-decomposition networks for cooperative multi-agent learning} {Value-decomposition networks for cooperative multi-agent learning}.{\BBCQ}
\newblock
\APACjournalVolNumPages{arXiv preprint arXiv:1706.05296}{}{}{,}
\newblock

\newblock

\PrintBackRefs{\CurrentBib}

\bibitem [\protect \citeauthoryear {%
Taki%
, Hasegawa%
\BCBL {}\ \BBA {} Fukumura%
}{%
Taki%
\ \protect \BOthers {.}}{%
{\protect \APACyear {1996}}%
}]{%
taki1996development}
\APACinsertmetastar {%
taki1996development}%
\begin{APACrefauthors}%
Taki, T.%
, Hasegawa, J\BHBI i.%
\BCBL {} Fukumura, T.%
\end{APACrefauthors}%
\unskip\
\newblock
\APACrefYearMonthDay{1996}{}{}.
\newblock
{\BBOQ}\APACrefatitle {Development of motion analysis system for quantitative evaluation of teamwork in soccer games} {Development of motion analysis system for quantitative evaluation of teamwork in soccer games}.{\BBCQ}
\newblock
 \APACrefbtitle {Proceedings of 3rd IEEE International Conference on Image Processing} {Proceedings of 3rd ieee international conference on image processing}\ (\BVOL~3, \BPGS\ 815--818).
\PrintBackRefs{\CurrentBib}

\bibitem [\protect \citeauthoryear {%
Teranishi%
, Tsutsui%
, Takeda%
\BCBL {}\ \BBA {} Fujii%
}{%
Teranishi%
\ \protect \BOthers {.}}{%
{\protect \APACyear {2022}}%
}]{%
teranishi2022evaluation}
\APACinsertmetastar {%
teranishi2022evaluation}%
\begin{APACrefauthors}%
Teranishi, M.%
, Tsutsui, K.%
, Takeda, K.%
\BCBL {} Fujii, K.%
\end{APACrefauthors}%
\unskip\
\newblock
\APACrefYearMonthDay{2022}{}{}.
\newblock
{\BBOQ}\APACrefatitle {Evaluation of creating scoring opportunities for teammates in soccer via trajectory prediction} {Evaluation of creating scoring opportunities for teammates in soccer via trajectory prediction}.{\BBCQ}
\newblock
 \APACrefbtitle {International Workshop on Machine Learning and Data Mining for Sports Analytics.} {International workshop on machine learning and data mining for sports analytics.}
\PrintBackRefs{\CurrentBib}

\bibitem [\protect \citeauthoryear {%
Toda%
, Teranishi%
, Kushiro%
\BCBL {}\ \BBA {} Fujii%
}{%
Toda%
\ \protect \BOthers {.}}{%
{\protect \APACyear {2022}}%
}]{%
toda2022evaluation}
\APACinsertmetastar {%
toda2022evaluation}%
\begin{APACrefauthors}%
Toda, K.%
, Teranishi, M.%
, Kushiro, K.%
\BCBL {} Fujii, K.%
\end{APACrefauthors}%
\unskip\
\newblock
\APACrefYearMonthDay{2022}{}{}.
\newblock
{\BBOQ}\APACrefatitle {Evaluation of soccer team defense based on prediction models of ball recovery and being attacked: A pilot study} {Evaluation of soccer team defense based on prediction models of ball recovery and being attacked: A pilot study}.{\BBCQ}
\newblock
\APACjournalVolNumPages{Plos one}{17}{1}{e0263051,}
\newblock

\newblock

\PrintBackRefs{\CurrentBib}

\bibitem [\protect \citeauthoryear {%
Umemoto%
, Tsutsui%
\BCBL {}\ \BBA {} Fujii%
}{%
Umemoto%
\ \protect \BOthers {.}}{%
{\protect \APACyear {2022}}%
}]{%
umemoto2022location}
\APACinsertmetastar {%
umemoto2022location}%
\begin{APACrefauthors}%
Umemoto, R.%
, Tsutsui, K.%
\BCBL {} Fujii, K.%
\end{APACrefauthors}%
\unskip\
\newblock
\APACrefYearMonthDay{2022}{}{}.
\newblock
{\BBOQ}\APACrefatitle {Location analysis of players in UEFA EURO 2020 and 2022 using generalized valuation of defense by estimating probabilities} {Location analysis of players in uefa euro 2020 and 2022 using generalized valuation of defense by estimating probabilities}.{\BBCQ}
\newblock
\APACjournalVolNumPages{arXiv preprint arXiv:2212.00021}{}{}{,}
\newblock

\newblock

\PrintBackRefs{\CurrentBib}

\bibitem [\protect \citeauthoryear {%
Wang%
\ \protect \BOthers {.}}{%
Wang%
\ \protect \BOthers {.}}{%
{\protect \APACyear {2024}}%
}]{%
wang2024tacticai}
\APACinsertmetastar {%
wang2024tacticai}%
\begin{APACrefauthors}%
Wang, Z.%
, Veli{\v{c}}kovi{\'c}, P.%
, Hennes, D.%
, Toma{\v{s}}ev, N.%
, Prince, L.%
, Kaisers, M.%
\BDBL {}others%
\end{APACrefauthors}%
\unskip\
\newblock
\APACrefYearMonthDay{2024}{}{}.
\newblock
{\BBOQ}\APACrefatitle {TacticAI: an AI assistant for football tactics} {Tacticai: an ai assistant for football tactics}.{\BBCQ}
\newblock
\APACjournalVolNumPages{Nature communications}{15}{1}{1906,}
\newblock

\newblock

\PrintBackRefs{\CurrentBib}

\bibitem [\protect \citeauthoryear {%
Yeung%
, Ide%
, Someya%
\BCBL {}\ \BBA {} Fujii%
}{%
Yeung%
, Ide%
\BCBL {}\ \protect \BOthers {.}}{%
{\protect \APACyear {2025}}%
}]{%
yeung2025openstarlab}
\APACinsertmetastar {%
yeung2025openstarlab}%
\begin{APACrefauthors}%
Yeung, C.%
, Ide, K.%
, Someya, T.%
\BCBL {} Fujii, K.%
\end{APACrefauthors}%
\unskip\
\newblock
\APACrefYearMonthDay{2025}{}{}.
\newblock
{\BBOQ}\APACrefatitle {OpenSTARLab: open approach for spatio-temporal agent data analysis in soccer} {Openstarlab: open approach for spatio-temporal agent data analysis in soccer}.{\BBCQ}
\newblock
\APACjournalVolNumPages{Complex \& Intelligent Systems}{11}{8}{342,}
\newblock

\newblock

\PrintBackRefs{\CurrentBib}

\bibitem [\protect \citeauthoryear {%
Yeung%
, Sit%
\BCBL {}\ \BBA {} Fujii%
}{%
Yeung%
, Sit%
\BCBL {}\ \BBA {} Fujii%
}{%
{\protect \APACyear {2025}}%
}]{%
yeung2023transformer}
\APACinsertmetastar {%
yeung2023transformer}%
\begin{APACrefauthors}%
Yeung, C.%
, Sit, T.%
\BCBL {} Fujii, K.%
\end{APACrefauthors}%
\unskip\
\newblock
\APACrefYearMonthDay{2025}{}{}.
\newblock
{\BBOQ}\APACrefatitle {Transformer-based neural marked spatio temporal point process model for analyzing football match events} {Transformer-based neural marked spatio temporal point process model for analyzing football match events}.{\BBCQ}
\newblock
\APACjournalVolNumPages{Applied Intelligence}{55}{5}{335,}
\newblock

\newblock

\PrintBackRefs{\CurrentBib}

\bibitem [\protect \citeauthoryear {%
Zahavy%
, Haroush%
, Merlis%
, Mankowitz%
\BCBL {}\ \BBA {} Mannor%
}{%
Zahavy%
\ \protect \BOthers {.}}{%
{\protect \APACyear {2018}}%
}]{%
zahavy2018learn}
\APACinsertmetastar {%
zahavy2018learn}%
\begin{APACrefauthors}%
Zahavy, T.%
, Haroush, M.%
, Merlis, N.%
, Mankowitz, D.J.%
\BCBL {} Mannor, S.%
\end{APACrefauthors}%
\unskip\
\newblock
\APACrefYearMonthDay{2018}{}{}.
\newblock
{\BBOQ}\APACrefatitle {Learn what not to learn: Action elimination with deep reinforcement learning} {Learn what not to learn: Action elimination with deep reinforcement learning}.{\BBCQ}
\newblock
\APACjournalVolNumPages{Advances in neural information processing systems}{31}{}{,}
\newblock

\newblock

\PrintBackRefs{\CurrentBib}

\end{thebibliography}

\end{document}